\pdfoutput=1

\documentclass[11pt]{article}

\usepackage{acl}
\usepackage{times}
\usepackage{latexsym}
\usepackage{subfigure}
\usepackage{booktabs} 
\usepackage{hyperref}
\usepackage{amsmath}
\usepackage{amssymb}
\usepackage{multirow}
\usepackage{array}
\usepackage{mathtools}
\usepackage{amsthm}
\usepackage{bm}
\usepackage{colortbl}
\usepackage{makecell}
\usepackage{textcomp}
\usepackage[T1]{fontenc}
\usepackage[utf8]{inputenc}
\usepackage{microtype}
\usepackage{inconsolata}
\usepackage{graphicx}
\usepackage{enumerate}
\usepackage{enumitem}
\usepackage[capitalize,noabbrev]{cleveref}
\usepackage[textsize=tiny]{todonotes}  
\usepackage{tikz}
\usetikzlibrary{calc}
\usepackage{etoolbox}

\newcommand*{\affaddr}[1]{#1} 
\newcommand*{\affmark}[1][*]{\textsuperscript{#1}}
\DeclareRobustCommand{\dashedbox}[1]{
    \tikz[baseline=(word.base)]{
        \node[inner sep=1pt, outer sep=0] (word) {#1};
        \draw[red, densely dashed, line width=1pt] 
            ($(word.south west) + (-1pt, -1pt)$) 
            rectangle 
            ($(word.north east) + (1pt, 1pt)$);
    }
}
\usepackage{fontawesome5}

\title{Towards Economical Inference: Enabling \\DeepSeek's Multi-Head Latent Attention in Any Transformer-based LLMs}

\author{
Tao Ji\textsuperscript{\rm 1,6},
Bin Guo\textsuperscript{\rm 2},
Yuanbin Wu\textsuperscript{\rm 2}, \\
\bf Qipeng Guo\textsuperscript{\rm 8},
Lixing Shen\textsuperscript{\rm 7},
Zhan Chen\textsuperscript{\rm 7},
Xipeng Qiu\textsuperscript{\rm 1},
Qi Zhang\textsuperscript{\rm 1,4,5},
Tao Gui\textsuperscript{\rm 3,4,5} \\
\affaddr{\affmark[1]School of Computer Science, Fudan University} \\
\affaddr{\affmark[2]School of Computer Science and Technology, East China Normal University}\\
\affaddr{\affmark[3]Institute of Modern Languages and Linguistics, Fudan University}\\
\affaddr{\affmark[4]Institute of Trustworthy Embodied Artificial Intelligence, Fudan University}\\
\affaddr{\affmark[5]Shanghai Collaborative Innovation Center of Intelligent Visual Computing}\\
\affaddr{\affmark[6]Pengcheng Laboratory}
\affaddr{\affmark[7]Hikvision Inc}
\affaddr{\affmark[8]Shanghai Al Lab}
\\
 \small{
    \href{mailto:taoji@fudan.edu.cn, tgui@fudan.edu.cn}{\{taoji,~tgui\}@fudan.edu.cn} \quad\quad
    \href{mailto:binguo@stu.ecnu.edu.cn, ybwu@cs.ecnu.edu.cn}{\{binguo@stu,~ybwu@cs\}.ecnu.edu.cn}
   }
}

\begin{document}
\maketitle

\begin{abstract}
    Multi-head Latent Attention (MLA) is an innovative architecture proposed by DeepSeek, 
designed to ensure efficient and economical inference by significantly compressing the Key-Value (KV) cache into a latent vector.
Compared to MLA, standard LLMs employing Multi-Head Attention (MHA) and its variants such as Grouped-Query Attention (GQA) exhibit significant cost disadvantages.
Enabling well-trained LLMs (e.g., Llama) to rapidly adapt to MLA without pre-training from scratch is both meaningful and challenging.
This paper proposes the first data-efficient fine-tuning method for transitioning from MHA to MLA (\textbf{MHA2MLA}), which includes two key components: 
for \textit{partial-RoPE}, we remove RoPE from dimensions of queries and keys that contribute less to the attention scores,  
for \textit{low-rank approximation}, we introduce joint SVD approximations based on the pre-trained parameters of keys and values. 
These carefully designed strategies enable MHA2MLA to recover performance using only a small fraction (0.6\%~to~1\%) of the data, significantly reducing inference costs while seamlessly integrating with compression techniques such as KV cache quantization. For example, the KV cache size of Llama2-7B is reduced by 92.19\%, with only a 1\% drop in LongBench performance.\footnote{
Our source code is publicly available at \url{https://github.com/JT-Ushio/MHA2MLA}.
}
\end{abstract}

\section{Introduction}
\label{sec:intro}

\begin{figure*}[t]
  \centering
  \includegraphics[width=0.9\linewidth]{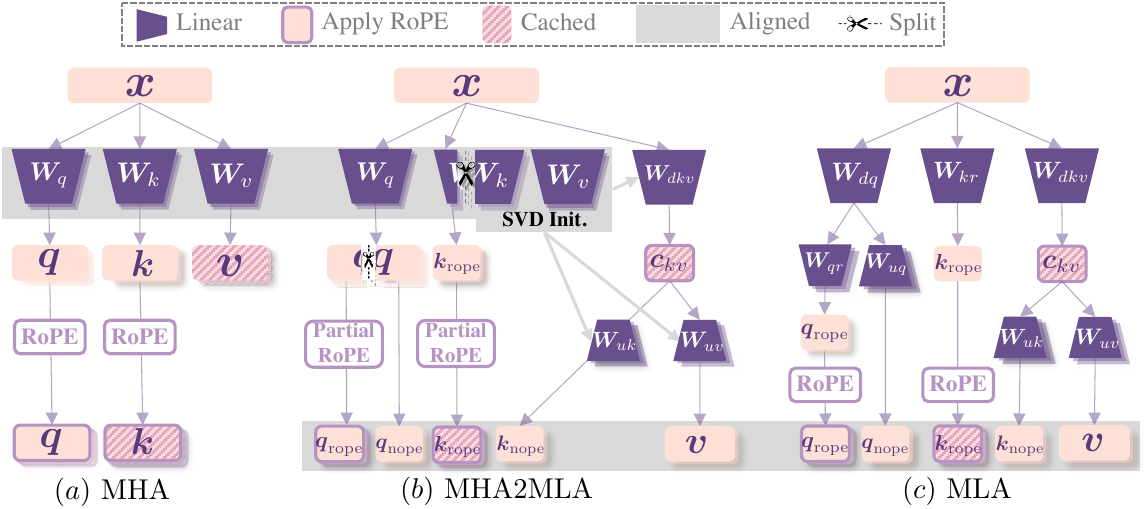}
  \caption{The diagram illustrates the MHA, MLA, and our MHA2MLA. It can be seen that the ``cached'' part is fully aligned with MLA after MHA2MLA. The input to the attention module is also completely aligned with MLA (the \colorbox{lightgray!60}{aligned region below}). Meanwhile, the parameters in MHA2MLA maximize the use of pre-trained parameters from MHA (the \colorbox{lightgray!60}{aligned region above}).}
  \vspace{-0.4cm}
  \label{fig:overview}
\end{figure*}



The rapid advancement of large language models (LLMs) has significantly accelerated progress toward artificial general intelligence (AGI), with model capabilities scaling predictably with parameter counts \cite{corr/abs-2001-08361}. 
However, these gains come at a steep cost: escalating computational demands for training and degraded inference throughput, resulting in substantial energy consumption and carbon emissions \cite{strubell-etal-2019-energy}. 


As downstream tasks grow increasingly complex, long-context processing and computationally intensive inference have become central to LLM applications
\cite{acl/AnG0ZLZKQ24}. 
A key bottleneck lies in the memory footprint of the Key-Value (KV) cache inherent to the Multi-Head Attention (MHA, \citeyear{nips/VaswaniSPUJGKP17}) mechanism, which scales linearly with sequence length and model size. 
To mitigate this, variants like Grouped-Query Attention (GQA, \citeyear{emnlp/AinslieLJZLS23})  and Multi-Query Attention (MQA, \citeyear{corr/abs-1911-02150})  have been explored. 
However, these methods reduce not only the KV cache size but also the number of parameters in the attention, leading to performance degradation.
The DeepSeek introduces Multi-Head Latent Attention (MLA, \citeyear{corr/abs-2405-04434}), an attention mechanism equipped with low-rank key-value joint compression. 
Empirically, MLA achieves superior performance compared with MHA, and meanwhile significantly reduces the KV cache during inference, thus boosting the inference efficiency. 


A critical yet unexplored question arises: \textbf{Can LLMs originally well-trained for MHA be adapted to enabling MLA for inference?} 
The inherent architectural disparities between MHA and MLA render zero-shot transfer impractical, while the prohibitive cost of pretraining from scratch makes this transition both technically challenging and underexplored in existing research.
To address this gap, we propose the first carefully designed MHA2MLA framework that maximizes parameter reuse from pre-trained MHA networks while aligning the KV cache storage and inference process with MLA's paradigm (\Cref{fig:overview}). Our framework features two pivotal technical innovations: partial rotary position embedding (partial RoPE) and low-rank approximation.
The primary objective of MHA2MLA is to achieve data-efficient performance recovery - restoring architecture-induced capability degradation using minimal fine-tuning data.


The inherent incompatibility between MLA's inference acceleration mechanism and RoPE necessitates architectural compromises. 
DeepSeek's solution preserves PEs in limited dimensions while compressing others, requiring strategic removal of RoPE dimensions (converting them to NoPE) in MHA to achieve MLA alignment. 
While higher removal ratios enhance compression efficiency, they exacerbate performance degradation, creating an efficiency-capability trade-off. 
Through systematically exploring RoPE removal strategies, we identify that contribution-aware dimension selection (retaining top-k dimensions ranked by attention score impact) optimally balances these competing objectives. 
Although previous studies have investigated training partial-RoPE LLMs from scratch~\cite{gpt-neo,corr/abs-2410-06205}, our work pioneers data-efficient fine-tuning for full-to-partial RoPE conversion in LLMs.



MLA reduces memory footprint by projecting keys and values into a low-rank latent representation space (stored in the KV cache). 
MHA2MLA can also apply low-rank approximation to the values and keys stripped of RoPE (NoPE dimensions). By performing  Singular Value Decomposition (SVD) on the pre-trained parameter matrices \( \bm{W}_v \) and \( \bm{W}_k \) corresponding to the NoPE subspaces, we compress these components into a latent space while maximizing the retention of knowledge learned by the original model.



Our main contributions are:
\begin{itemize}[leftmargin=*,itemsep=0pt, topsep=0pt, parsep=0pt]
    \item we introduce MHA2MLA, the first parameter-efficient fine-tuning framework that adapts pre-trained MHA-based LLMs to the MLA architecture using only 0.6\%~to~1\% of training data without training from scratch.
    \item we demonstrate that the MHA2MLA architecture can be integrated with KV-cache quantization to achieve more economical inference (up to a 96.87\% reduction).
    \item we conduct experiments across five model sizes (from 135M to 13B, covering both MHA and GQA), and detailed ablation studies to provide guidance and insights for MHA2MLA.
\end{itemize}


\section{Preliminary}
\label{sec:preliminary}

\subsection{Multi-Head Attention (MHA)}  
Given an input sequence \(\{\bm{x}_1,\dots, \bm{x}_l\} \in \mathbb{R}^{l \times d}\), standard MHA \cite{nips/VaswaniSPUJGKP17} projects each token \(\bm{x}_i\) into queries \(\bm{q}_i^{(h)} = \bm{x}_i \bm{W}_q^{(h)}\), keys \(\bm{k}_i^{(h)} = \bm{x}_i \bm{W}_k^{(h)}\), and values \(\bm{v}_i^{(h)} = \bm{x}_i \bm{W}_v^{(h)}\), where \( \bm{W}_q^{(h)}, \bm{W}_k^{(h)}, \bm{W}_v^{(h)} \in \mathbb{R}^{d \times d_h} \) for each head \( h \in \{1, \dots, n_h\} \). The Rotary positional encoding (RoPE, \citeyear{journals/ijon/SuALPBL24}) is applied to queries and keys (e.g., $\bm{k}_{i,\text{rope}}^{(h)}=\text{RoPE}(\bm{k}_{i}^{(h)})$), followed by scaled dot-product attention\footnote{
We ignore here the $\frac{1}{\sqrt{d}}$ scaling factor for ease of notation.
}:  
\begin{align}    
    &\bm{o}_i^{(h)} = \text{Softmax}\left(\bm{q}_{i,\text{rope}}^{(h)} \bm{k}_{\le i,\text{rope}}^{(h)\top}\right) \bm{v}_{\le i}^{(h)}, \nonumber \\
    &\text{MHA}(\bm{x}_i) = \left[\bm{o}_i^{(1)}, \dots, \bm{o}_i^{(n_h)}\right] \bm{W}_o, 
\end{align}
where $\bm{W}_o \in \mathbb{R}^{(n_h d_h) \times d}$ and $[\cdot,\cdot]$ means vector concatenate.
During autoregressive inference, MHA stores the KV cache \(\{\bm{k}_{\text{rope}}^{(h)}, \bm{v}^{(h)}\}_{h=1}^{n_h}\) of size \( \mathcal{O}(2 l n_h d_h) \), growing linearly with sequence length \( l \), posing memory bottlenecks.  

\paragraph{Variants:}  
Grouped-Query Attention (GQA, \citeyear{emnlp/AinslieLJZLS23}) shares keys/values across \( n_g \) groups (\( n_g \ll n_h \)) to reduce the KV cache. For each head \( h \), it maps to group \( g=\lfloor \frac{h\times n_g}{n_h} \rfloor \):  
\begin{align}
    &\bm{o}_i^{(h)} = \text{Softmax}\left(\bm{q}_{i,\text{rope}}^{(h)} \bm{k}_{\le i,\text{rope}}^{(g)\top}\right) \bm{v}_{\le i}^{(g)}, \nonumber \\
    &\text{GQA}(\bm{x}_i) = \left[\bm{o}_i^{(1)}, \dots, \bm{o}_i^{(n_h)}\right] \bm{W}_o.
\end{align}
Multi-Query Attention (MQA, \citeyear{strubell-etal-2019-energy}) is a special case of GQA with \( n_g = 1 \), i.e., all heads share a single global key/value. 
While reducing the KV cache to \( \mathcal{O}(2 l n_g d_h) \), these methods degrade performance due to parameter pruning.

\subsection{Multi-Head Latent Attention (MLA)}  

MLA~\cite{corr/abs-2405-04434} introduces a hybrid architecture that decouples PE from latent KV compression. 
For each head $h$, the input $\bm{x}_i$ is projected into two complementary components:  

\paragraph{Position-Aware Component}
A subset of dimensions retains PE to preserve positional sensitivity:  
$$
 \bm{q}_{i,\text{rope}}^{(h)}, \bm{k}_{i,\text{rope}} = \text{RoPE}
    \left( 
        \bm{x}_i \bm{W}_{dq}
        \bm{W}_{qr}^{(h)}, \bm{x}_i \bm{W}_{kr} 
    \right),
$$
where $\bm{W}_{dq} \in \mathbb{R}^{d \times d_q}$, $\bm{W}_{qr}^{(h)} \in \mathbb{R}^{d_q\times d_r}$, $\bm{W}_{kr} \in \mathbb{R}^{d \times d_r}$ project queries/keys into a RoPE-preserved component of dimension $d_r$.  

\paragraph{Position-Agnostic Component}
The remaining dimensions $d_c$ are stripped of PE (i.e., NoPE), $\bm{k}_{i,\text{nope}}^{(h)}$ and $\bm{v}_{i}^{(h)}$ and compressed into a shared latent vector $\bm{c}_{i,kv}^{(h)}$:
\begin{align}
   \bm{q}_{i,\text{nope}}^{(h)} &= \bm{x}_i \bm{W}_{dq} \bm{W}_{qc}^{(h)},\nonumber\\ \bm{c}_{i,kv} &= \bm{x}_i \bm{W}_{dkv}, \nonumber\\
   \bm{k}_{i,\text{nope}}^{(h)}, \bm{v}_{i}^{(h)} &= \bm{c}_{i,kv} \bm{W}_{uk}^{(h)}, \bm{c}_{i,kv} \bm{W}_{uv}^{(h)} \nonumber,
\end{align}
where $\bm{W}_{qc}^{(h)} \in \mathbb{R}^{d_q \times d_c}$, 
$\bm{W}_{dkv} \in \mathbb{R}^{d \times d_{kv}}$, 
$\bm{W}_{uk}^{(h)} \in \mathbb{R}^{d_{kv} \times d_c}$, 
$\bm{W}_{uv}^{(h)} \in \mathbb{R}^{d_{kv} \times d_h}$.
Note that $d_r+d_c = d_h$.
The attention output of MLA combines both components:  
\begin{align}    
    &\bm{o}_i^{(h)}=\text{Softm}\text{ax}\left(\bm{q}_{i,\text{rope}}^{(h)} \bm{k}_{\le i,\text{rope}}^{(h)\top}+\bm{q}_{i,\text{nope}} \bm{k}_{\le i,\text{nope}}^{(h)\top}\right) \nonumber\\
    &\quad \quad \quad \cdot \bm{v}_{\le i}^{(h)} \nonumber \\
    &\quad \quad \text{MLA}(\bm{x}_i) = \left[\bm{o}_i^{(1)}, \dots, \bm{o}_i^{(n_h)}\right] \cdot \bm{W}_o.
\end{align}
Unlike MHA and its variants, MLA stores the latent vector $\bm{c}_{kv}$ and $\bm{k}_{i,\text{rope}}^{(h)}$ (\( \mathcal{O}\left(ld_r+ld_{kv})\right) \)) instead of full-rank \( \bm{k}_i, \bm{v}_i \) (\( \mathcal{O}(2ln_hd_h) \)), where \( (d_r+d_{kv}) \ll 2n_hd_h \). 

\paragraph{Why does MLA need to separate RoPE and NoPE?}
MLA introduces matrix merging techniques for the NoPE portion during inference, 
effectively reducing memory usage. 
For the dot product operation \(\bm{q}_{i,\text{nope}}^{(h)} \bm{k}_{j,\text{nope}}^{(h)\top}\), the following identity transformation can be applied
\footnote{To simplify the notation, we omit the superscript $^{(h)}$. 
Matrices $\bm{W}_{uv}$ and $\bm{W}_{o}$ can also be merged, please refer to Appendix C by \citet{corr/abs-2405-04434}.}:
\begin{align}    
    \bm{q}_{i,\text{nope}} \bm{k}_{j,\text{nope}}^{\top} 
    &= \left(\bm{x}_i \bm{W}_{dq} \bm{W}_{qc}\right)\left(\bm{c}_{j,kv} \bm{W}_{uk}\right)^\top \nonumber\\
    &= \bm{x}_i \left(\bm{W}_{dq} \bm{W}_{qc}\bm{W}_{uk}^\top\right)\bm{c}_{j,kv}^\top \nonumber
\end{align}
where $\left(\bm{W}_{dq} \bm{W}_{qc}\bm{W}_{uk}^\top\right)$ can be pre-merged into a single matrix, and $\bm{c}_{j,kv}$ is already stored in the KV cache. 
As for the RoPE portion, the RoPE($\cdot$) function multiplies the input vector by the rotation matrix (e.g., RoPE($\bm{q}_i$) = $\bm{q}_i\bm{R}_i$, $\bm{R}_i$'s specific form will be introduced in \Cref{ssec:partial_rope}). 
Therefore, the identity transformation becomes as follows:
\begin{align}    
    \bm{q}_{i,\text{rope}} \bm{k}_{j,\text{rope}}^{\top} 
    &= \left(\bm{x}_i \bm{W}_{dq} \bm{W}_{qr} \bm{R}_i\right)\left(\bm{x}_j \bm{W}_{kr} \bm{R}_j\right)^\top \nonumber\\
    &= \bm{x}_i \left(\bm{W}_{dq} \bm{W}_{qc}\bm{R}_{j-i}\bm{W}_{kr}^\top\right)\bm{x}_{j}^\top \nonumber
\end{align}
Since \(\left(\bm{W}_{dq} \bm{W}_{qc} \bm{R}_{j-i} \bm{W}_{kr}^\top\right)\) is related to the relative position \(j-i\), it cannot be merged into a fixed matrix. 
Considering that the relative distances in LLMs can be very long, such as 128K, the RoPE portion is better suited to be computed using the original form.

\section{MHA2MLA}

\subsection{Partial-RoPE}
\label{ssec:partial_rope}

To enable migration from standard MHA to MLA, we propose partial-RoPE finetuning, a strategy that removes RoPE from a targeted proportion of dimensions and converts them into NoPE. 
Critically, while prior work has explored training LLMs with partial-RoPE from scratch (achieving marginally better perplexity than full-RoPE \cite{gpt-neo,corr/abs-2410-06205}), no existing method addresses how to efficiently adapt pre-trained full-RoPE models (e.g., Llama) to partial-RoPE without costly retraining. 
Our work bridges this gap by systematically evaluating partial-RoPE variants to identify the most data-efficient fine-tuning protocol for recovering model performance post-adaptation.

\begin{figure}[t]
  \centering
  \includegraphics[width=\linewidth]{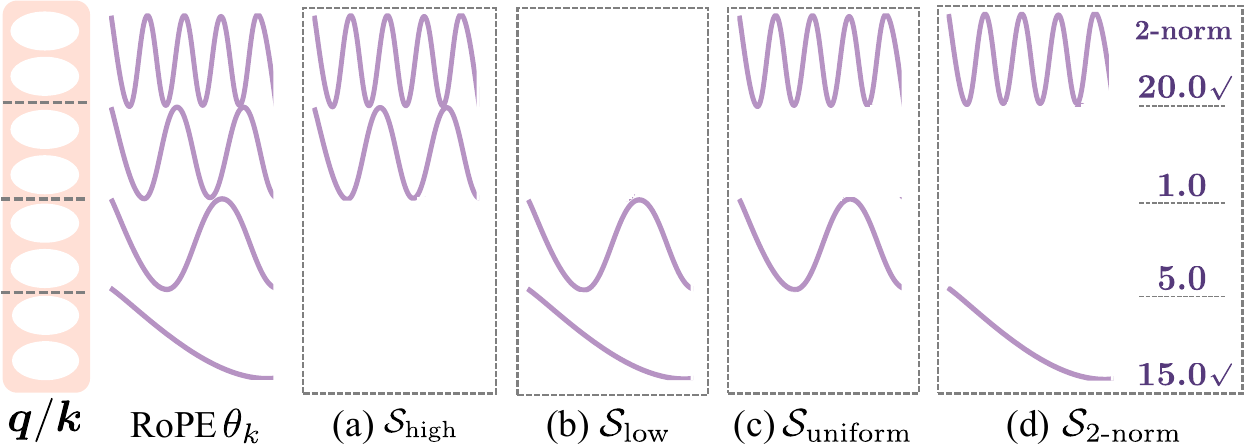}
  \caption{Illustration of $\mathcal{S}_{\text{high}}$, $\mathcal{S}_{\text{low}}$, $\mathcal{S}_{\text{uniform}}$, $\mathcal{S}_{\text{2-norm}}$. Where $d_h=8$ and $r=2$.}
  %
  \vspace{-0.2cm}
  \label{fig:partial_rope}
\end{figure}

\begin{figure}[t]
  \centering
  \includegraphics[width=\linewidth]{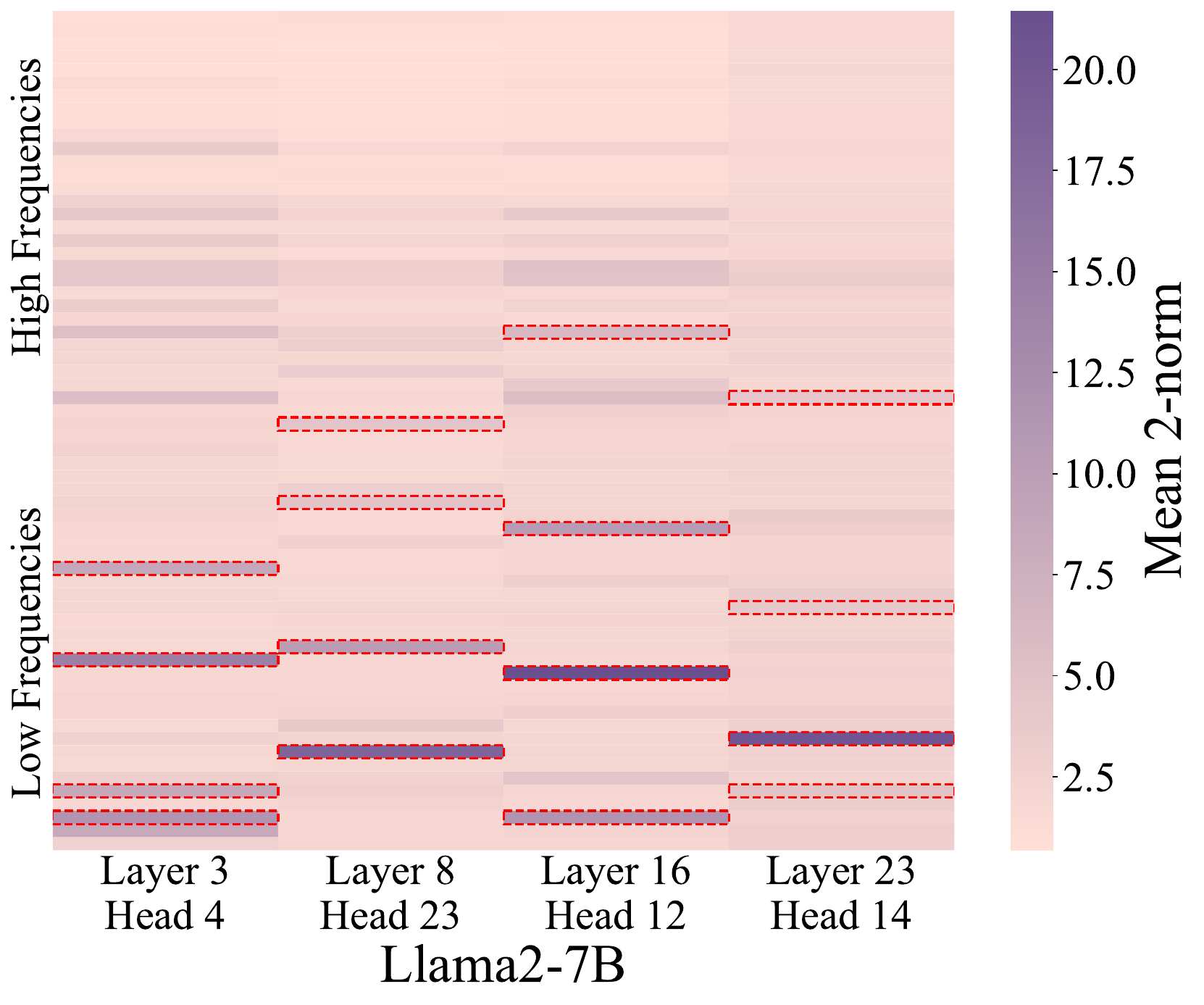}
  \caption{Visualization of Head-wise 2-norm Contribution for Llama2-7B. We randomly selected 4 heads, and the \dashedbox{red dashed box} highlights the top-$4$ frequency subspaces chosen when $r=4$. It can be seen that different heads tend to focus on different frequency subspaces, which validates the rationality of our $\mathcal{S}_{\text{2-norm}}$ method.}
  \label{fig:7b_2norm}
\end{figure}


\paragraph{MHA's Full-RoPE} encodes positional information into queries and keys through frequency-specific rotations. Formally, given a query vector \(\bm{q}_i \in \mathbb{R}^{d_h}\) and key vector \(\bm{k}_i \in \mathbb{R}^{d_h}\), we partition them into 2D chunks:  
\begin{align*}
\bm{q}_i,\bm{k}_i = \left[\bm{q}_i^{[2k, 2k+1]}\right]_{0 \leq k < \frac{d_h}{2}}, \left[\bm{k}_i^{[2k, 2k+1]}\right]_{0 \leq k < \frac{d_h}{2}},
\end{align*}
where \(\bm{q}_i^{[2k, 2k+1]} \in \mathbb{R}^2\) denotes the \(k\)-th 2D subspace. Each chunk undergoes a rotation by position-dependent angles \(\theta_k = \beta^{-2k/{d_h}}\), forming a spectrum of wavelengths.
High-frequency components, e.g., $k=0$, rotate rapidly at 1 radian per token.  
Low-frequency components, e.g., $k=\frac{d_h}{2}-1$, rotate slowly at \(\sim \beta^{1/d_h}\) radians per token.  
The base wavelength \(\beta\), typically set to \(10^4\) \cite{journals/ijon/SuALPBL24} or \(5\!\times\!10^5\).


Formally, for each 2D chunk \( \bm{q}_i^{[2k, 2k+1]} \) and \( \bm{k}_i^{[2k, 2k+1]} \), the rotation matrix at position \( i \) is defined as:  
\[
\bm{R}_i^{[2k, 2k+1]}(\theta_k) = 
\begin{bmatrix} 
\cos(i\theta_k) & -\sin(i\theta_k) \\ 
\sin(i\theta_k) & \cos(i\theta_k) 
\end{bmatrix}.
\]  
Thus, applying RoPE to queries and keys becomes:  
\begin{align*}
\bm{q}_{i, rope} = \left[\bm{R}_i^{[2k, 2k+1]}(\theta_k)\bm{q}_i^{[2k, 2k+1]}\right]_{0 \leq k < \frac{d_h}{2}},  \\
\bm{k}_{i, rope} = \left[\bm{R}_i^{[2k, 2k+1]}(\theta_k)\bm{k}_i^{[2k, 2k+1]}\right]_{0 \leq k < \frac{d_h}{2}}.
\end{align*}



\paragraph{Full-RoPE to Partial-RoPE Strategies}  
Given $r$ retained rotational subspaces($r=\frac{d_r}{2}\ll$ total subspaces $\frac{d_h}{2}$, we propose four strategies (illustrated in \Cref{fig:partial_rope}) to select which \( r \) subspaces preserve RoPE encoding:

 \textbf{High-Frequency Preservation} retain the \( r \) fastest-rotating (high-frequency) subspaces:
    \[
    \mathcal{S}_{\text{high}} = \left\{ k \,\vert\, 0 \leq k < r \right\}.
    \]
It is consistent with the p-RoPE method proposed in \citet{corr/abs-2410-06205}, where they explored settings in which \(r\) constituted 25\%, 50\%, and 75\% of the total subspaces, and observed a slight advantage over full-RoPE in LLMs trained from scratch.
    
 \textbf{Low-Frequency Preservation} retain the \( r \) slowest-rotating (low-frequency) subspaces:
\[
\mathcal{S}_{\text{low}} = \left\{ k \,\big|\, \frac{d_h}{2} - r \leq k < \frac{d_h}{2} \right\}.
\]
It was chosen as a controlled experiment for the high-frequency strategy.
    
 \textbf{Uniform Sampling} select \( r \) subspaces with equidistant intervals:
\[
\mathcal{S}_{\text{uniform}} = \left\{ \left\lfloor k \frac{d_h}{2r} \right\rfloor \,\bigg|\, 0 \leq k < r \right\}
\]
This balances high- and low-frequency components through geometric spacing.
In practice, \(2r\) typically divides \(d_h\).
It is similar to the partial RoPE used in GPT-Neo~\cite{gpt-neo}.
    
\textbf{Head-wise 2-norm Contribution} 
\citet{corr/abs-2410-06205} were the first to propose the 2-norm contribution to investigate whether these frequencies are utilized and how they are helpful. 
This approach is based on the observation that, according to the Cauchy-Schwarz inequality, the influence of the \( k \)-th frequency subspace on the attention logits is upper-bounded by the 2-norm of the corresponding query and key components, i.e., $\left|\left\langle\mathbf{q}_i^{[2k,2k+1]}, \mathbf{k}_j^{[2k,2k+1]}\right\rangle\right| \leqslant\left\|\mathbf{q}_i^{[2k,2k+1]}\right\|\left\|\mathbf{k}_j^{[2k,2k+1]}\right\|$.
For each head \( h \), we compute the mean 2-norm score for each subspace in an LLM over long sequences
\footnote{
The 2-norm calculation detail is placed in \Cref{app:2_norm}.
}.
Then, we propose to rank all subspaces by their 2-norm score and select the top-$r$:
\begin{align*}
    \mathcal{S}_{\text{2-norm}}\!=\!\underset{0\le k<\frac{d_h}{2}}{\text{top-}r} \left( \left\|\mathbf{q}_*^{[2k,2k+1]}\right\|\left\|\mathbf{k}_*^{[2k,2k+1]}\right\| \right).
\end{align*}
This head-specific selection adaptively preserves rotation-critical subspaces.
\Cref{fig:7b_2norm} visualizes the 2-norm of  Llama2-7B's four heads.

We will analyze the effectiveness of the four strategies in \Cref{ssec:ablation_study} and conduct an ablation study on the essential hyperparameter $r$ in \Cref{app:pe-dim}.
For all strategies, non-selected subspaces (\( k \notin \mathcal{S} \)) become NoPE dimensions, enabling seamless integration with MLA's latent compression.

\begin{figure}[t]
  \centering
  \includegraphics[width=0.85\linewidth]{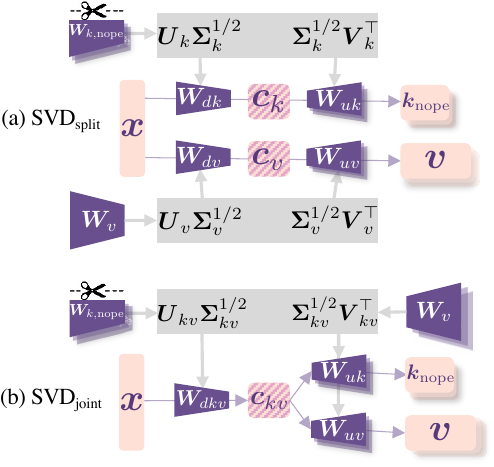}
  \caption{Illustration of \textbf{SVD\textsubscript{split}} and \textbf{SVD\textsubscript{joint}}. In the multi-head setting, we adhere to the standard MLA approach, performing SVD on the merged multi-heads rather than on each head individually (e.g., $\bm{U}_{kv} \in \mathbb{R}^{n_hd_h \times n_hd_{kv}}$.}
  \vspace{-0.4cm}
  %
  \label{fig:svd_overview}
\end{figure}

\subsection{Low-rank Approximation}
\label{sec:low_rank_appro}




After transitioning from full RoPE to partial RoPE, we obtain the first component of the KV cache in MLA, represented as: \(\bm{k}_{i, rope} = \left[\bm{R}_i^{[2k,2k+1]}(\theta_k)\bm{k}_i^{[2k,2k+1]}\right]_{k \in \mathcal{S}}\). 
Our next goal is to derive the second component, \(\bm{c}_{i,kv} \in \mathbb{R}^{d_{kv}}\), which serves as a low-rank representation of \(\bm{k}_{i,\text{nope}}\) and \(\bm{v}_i\).

Given the keys $\bm{k}_i = \bm{x}_i \bm{W}_k$ and values $\bm{v}_i = \bm{x}_i \bm{W}_v$ in MHA,
we first extract the subspace of $\bm{W}_k$ corresponding to \(\bm{k}_{i,\text{nope}}\), i.e., the dimensions not included in \(\mathcal{S}\), yielding: $\bm{k}_{i,\text{nope}} = \bm{x}_i \bm{W}_{k,\text{nope}}$.
We propose two Singular Value Decomposition (SVD)-based strategies (Illustrated in \Cref{fig:svd_overview}) to preserve pre-trained knowledge while achieving rank reduction:

\paragraph{Decoupled SVD (SVD\textsubscript{split})}  
Separately decompose \(\bm{W}_{k,\text{nope}}\) and \(\bm{W}_v\) into truncated SVDs, allocating \(d_{kv}/2\) dimensions to each:  
\[
\bm{W}_{k,\text{nope}} = \bm{U}_k \bm{\Sigma}_k \bm{V}_k^\top, \quad \bm{W}_v = \bm{U}_v \bm{\Sigma}_v \bm{V}_v^\top,
\]  
where \(\bm{U}_k, \bm{U}_v,\bm{V}_{k},\bm{V}_{v}\in \mathbb{R}^{d_h \times \frac{d_{kv}}{2}}\), \(\bm{\Sigma}_k, \bm{\Sigma}_v \in \mathbb{R}^{\frac{d_{kv}}{2} \times \frac{d_{kv}}{2}}\). The down-projection matrices $\bm{W}_{d*}$ and up-projection matrices $\bm{W}_{u*}$ become:  
\begin{gather*}
\bm{W}_{dk} = \bm{U}_k \bm{\Sigma}_k^{1/2}, \quad \bm{W}_{uk} = \bm{\Sigma}_k^{1/2} \bm{V}_k^\top, \\ 
\bm{W}_{dv} = \bm{U}_v \bm{\Sigma}_v^{1/2}, \quad \bm{W}_{uv} = \bm{\Sigma}_v^{1/2} \bm{V}_v^\top.
\end{gather*}
The low-rank representation \( \bm{c}_{i, kv} \) can be constructed using $\bm{c}_{i,kv} = \left[ \bm{x}_i \bm{W}_{dk}, \bm{x}_i \bm{W}_{dv} \right]$.

\paragraph{Joint SVD (SVD\textsubscript{joint})}  
To preserve interactions between \(\bm{K}_{\text{nope}}\) and \(\bm{V}\), we jointly factorize the concatenated matrix:  
\[
[\bm{W}_{k,\text{nope}}, \bm{W}_v] = \bm{U}_{kv} \bm{\Sigma}_{kv} \bm{V}_{kv}^\top,
\]  
where \(\bm{U}_{kv},\bm{V}_{kv} \in \mathbb{R}^{d_h \times d_{kv}}\), \(\bm{\Sigma}_{kv} \in \mathbb{R}^{d_{kv} \times d_{kv}}\). The latent projection is then:  
\begin{gather*}
\bm{W}_{dkv} = \bm{U}_{kv} \bm{\Sigma}_{kv}^{1/2}, \\ 
\bm{W}_{uk}\!=\!\bm{\Sigma}_{kv}^{1/2} \bm{V}_{kv}[:, :-d_v], \bm{W}_{uv}\!=\!\bm{\Sigma}_{kv}^{1/2} \bm{V}_{kv}[:, d_v:].    
\end{gather*}
This jointly optimizes the latent space for both keys and values, i.e., $\bm{c}_{i,kv} = \bm{x}_i \bm{W}_{dkv}$, retaining cross-parameter dependencies critical for autoregressive generation
\footnote{
We describe the economical inference process of MHA2MLA in \Cref{app:mha2mla_infer}.
}.  
\Cref{ssec:ablation_study} shows \textbf{SVD\textsubscript{joint}} outperforming \textbf{SVD\textsubscript{split}}, validating that joint factorization better preserves pre-trained knowledge.

\section{Experiment}
\label{sec:exper}

\begin{table*}[t]
\centering
\small
\begin{tabular}{l@{}lccr@{\hspace{2pt}}lcccccc}
  \toprule
  \multicolumn{2}{l}{\textbf{Model}} & \textbf{Tokens} & \textbf{KV Mem.} & \multicolumn{2}{c}{\textbf{Avg.}} & \textbf{MMLU} & \textbf{ARC} & \textbf{PIQA} & \textbf{HS} & \textbf{OBQA} & \textbf{WG}\\
  \midrule
  \rowcolor{gray!10} \multicolumn{2}{l}{135M$_{\text{SmolLM}}$} & 600B &   
  & \multicolumn{2}{l}{44.50} & 29.80 & 42.43 & 68.06 & 41.09 & 33.60 & 52.01\\
  \textit{- GQA}& ~$d_{kv}\!=\!128$ &  &   
  & 44.42 & & 29.91 & 41.71 & 68.28 & 41.33 & 33.80 & 51.46 \\
  \multirow{3}{*}{\textit{- GQA2MLA}}   & ~$d_{kv}\!=\!32$ & 6B & -68.75\%  
  & 43.21 &\textsubscript{-1.21} & 29.50 & 40.84 & 67.08 & 38.34 & 33.20 & 50.28 \\
  & ~$d_{kv}\!=\!16$ & (1\%) & -81.25\%  
  & 42.18 &\textsubscript{-2.24} & 28.79 & 40.11 & 65.94 & 36.68 & 31.20 & 50.36 \\
  & ~$d_{kv}\!=\!8$ &  & -87.50\% 
  & 41.04 &\textsubscript{-3.38} & 28.49 & 38.17 & 64.36 & 33.93 & 31.60 & 49.72 \\
  \midrule
  \rowcolor{gray!10} \multicolumn{2}{l}{360M$_{\text{SmolLM}}$} & 600B &   & 
  \multicolumn{2}{l}{49.60} & 33.70 & 49.82 & 71.87 & 51.65 & 37.60 & 52.96 \\
  \textit{- GQA}& ~$d_{kv}\!=\!128$ &  &  & 49.51 & & 34.08 & 49.89 & 71.60 & 51.67 & 37.20 & 52.64 \\
  \multirow{3}{*}{\textit{- GQA2MLA}} & ~$d_{kv}\!=\!32$ & 6B & -68.75\% & 48.14 & \textsubscript{-1.37} & 32.91 & 48.34 & 70.51 & 48.56 & 36.80 & 51.70 \\
  & ~$d_{kv}\!=\!16$ & (1\%) & -81.25\%  & 46.88 & \textsubscript{-2.63} & 31.85 & 46.07 & 70.62 & 46.48 & 35.80 & 50.43 \\
  & ~$d_{kv}\!=\!8$  & & -87.50\% & 45.84 & \textsubscript{-3.67} & 30.93 & 44.41 & 69.48 & 43.49 & 36.00 & 50.75 \\
  \midrule
 \rowcolor{gray!10} \multicolumn{2}{l}{1B7$_{\text{SmolLM}}$} & 1T &   
  & \multicolumn{2}{l}{55.90} & 39.27 & 59.87 & 75.73 & 62.93 & 42.80 & 54.85 \\
  \textit{- MHA}&~$d_{kv}\!=\!128$ & &   
  & 55.71 & & 38.66 & 59.02 & 75.79 & 62.60 & 43.20 & 55.01 \\
  \multirow{3}{*}{\textit{- MHA2MLA}} 
  & ~$d_{kv}\!=\!32$ & 6B & -68.75\%  
  & 54.66 &\textsubscript{-1.05}& 37.79 & 56.54 & 75.19 & 61.30 & 41.40 & 55.72 \\
  & ~$d_{kv}\!=\!16$ & (0.6\%) & -81.25\%  
  & 54.28 &\textsubscript{-1.43}& 37.79 & 56.33 & 75.68 & 60.59 & 41.00 & 54.30 \\
  & ~$d_{kv}\!=\!8$ & & -87.50\% 
  & 52.79 &\textsubscript{-2.92}& 36.69 & 54.29 & 74.54 & 58.49 & 39.20 & 53.51 \\
  \midrule  \rowcolor{gray!10}\multicolumn{2}{l}{7B$_{\text{Llama2}}$} & 2T &   
  & \multicolumn{2}{l}{59.85} & 41.43 & 59.24 & 78.40 & 73.29 & 41.80 & 64.96 \\
  \textit{- MHA}& ~$d_{kv}\!=\!256$ & &   
  & 59.50 & & 40.30 & 60.05 & 77.91 & 70.20 & 45.00 & 63.54 \\
  \multirow{3}{*}{\textit{- MHA2MLA}} 
  & ~$d_{kv}\!=\!64$ & 12B & -68.75\%  
  & 59.21 &\textsubscript{-0.29}& 40.90 & 59.53 & 77.26 & 70.05 & 45.20 & 62.51 \\
  & ~$d_{kv}\!=\!32$ & (0.6\%) & -81.25\%  
  & 59.20 &\textsubscript{-0.30}& 40.96 & 59.74 & 77.26 & 69.81 & 44.00 & 63.46 \\
  & ~$d_{kv}\!=\!16$ & & -87.50\%& 58.48&\textsubscript{-1.02} & 40.15 & 58.53 & 77.20 & 68.88 & 45.00 & 61.09 \\
  \midrule  \rowcolor{gray!10}\multicolumn{2}{l}{13B$_{\text{Llama2}}$} & 2T &   
  & \multicolumn{2}{l}{62.65} & 43.56 & 62.68 & 80.41 & 76.99 & 43.80 & 68.43 \\
  \textit{- MHA}& ~$d_{kv}\!=\!256$ & &   
  & 60.72 & & 42.89 & 61.75 & 77.69 & 72.22 & 44.00 & 65.75 \\
  \multirow{3}{*}{\textit{- MHA2MLA}} 
  & ~$d_{kv}\!=\!64$ & 12B & -68.75\%  
  & 60.45 &\textsubscript{-0.27}& 42.49 & 62.31 & 78.78 & 71.59 & 43.20 & 64.33 \\
  & ~$d_{kv}\!=\!32$ & (0.6\%) & -81.25\%  
  & 60.49 &\textsubscript{-0.23}&42.00 & 61.64 & 78.73 & 72.71 & 42.80 & 65.04 \\
  & ~$d_{kv}\!=\!16$ & & -87.50\%& 59.68&\textsubscript{-1.04} & 41.31 & 60.39 & 77.69 & 71.54 & 43.60 & 63.54
 \\
  \bottomrule
\end{tabular}
\caption{Commonsense reasoning ability of four LLMs with MHA2MLA or GQA2MLA. The six benchmarks include MMLU (\citeyear{MMLU}), ARC easy and challenge (ARC,~\citeyear{ARC}), PIQA (\citeyear{PIQA}), HellaSwag (HS,~\citeyear{HS}), OpenBookQA (OBQA,~\citeyear{OBQA}), Winogrande (WG,~\citeyear{WG}).}
\vspace{-0.4cm}
\label{tab:cs}
\end{table*}


We evaluate our method on LLMs of varying scales (SmolLM-135M/360M/1B7, Llama2-7B/13B) pre-trained with MHA or GQA.
We chose the SmolLM-series\footnote{\url{https://huggingface.co/collections/HuggingFaceTB/smollm-6695016cad7167254ce15966}}
because its pretraining data and framework are both open-source, which can minimize the gap in fine-tuning data and processes. 
We chose Llama2-7B\footnote{\url{https://huggingface.co/meta-llama/Llama-2-7b}} because it is one of the widely used open-source LLMs (but its pretraining data is not open-source, there is a potential gap in fine-tuning data).

We denote the architectural migration using MHA2MLA and GQA2MLA, respectively.\footnote{
The details of the fine-tuning process (including data and hyperparameters) are provided in \Cref{app:ft_details}.
}
Both adopt \textit{data-efficient full-parameter fine-tuning}, with the head-wise 2-norm selection ($\mathcal{S}_{\text{2-norm}}$, $r=\frac{d_h}{16}$) for Partial-RoPE and joint SVD factorization (\textbf{SVD\textsubscript{joint}}) for low-rank approximation as default configurations. 
Our experiments address three critical questions:  
\begin{enumerate}[leftmargin=*,itemsep=0pt, topsep=0pt, parsep=0pt]
    \item How does MHA2MLA minimize accuracy degradation induced by architectural shifts? 
    \item What does MHA2MLA achieve in the KV cache reduction ratio?  
    \item Can MHA2MLA integrate with KV cache quantization for compound gains?  
\end{enumerate}

\subsection{Commonsense Reasoning Tasks}  
\label{ssec:general_res}

\paragraph{Main Results}

As shown in \Cref{tab:cs}, our method achieves efficient architectural migration across five model scales (135M to 13B) under varying KV cache compression ratios (via latent dimension \( d_{kv} \)). 
First, when comparing the performance of our fine-tuning approach with the original LLM, we observe only minor changes in performance across the five base models: a -0.08\% decrease on the 135M, -0.09\% on the 360M, -0.19\% on the 1B7, -0.35\% on the 7B and -1.93\% on the 13B. 
This suggests that the fine-tuning data does not significantly degrade or improve the performance of the original model, providing an appropriate experimental setting for the MHA2MLA framework.

\begin{figure}[t]
  \centering
  \includegraphics[width=\linewidth]{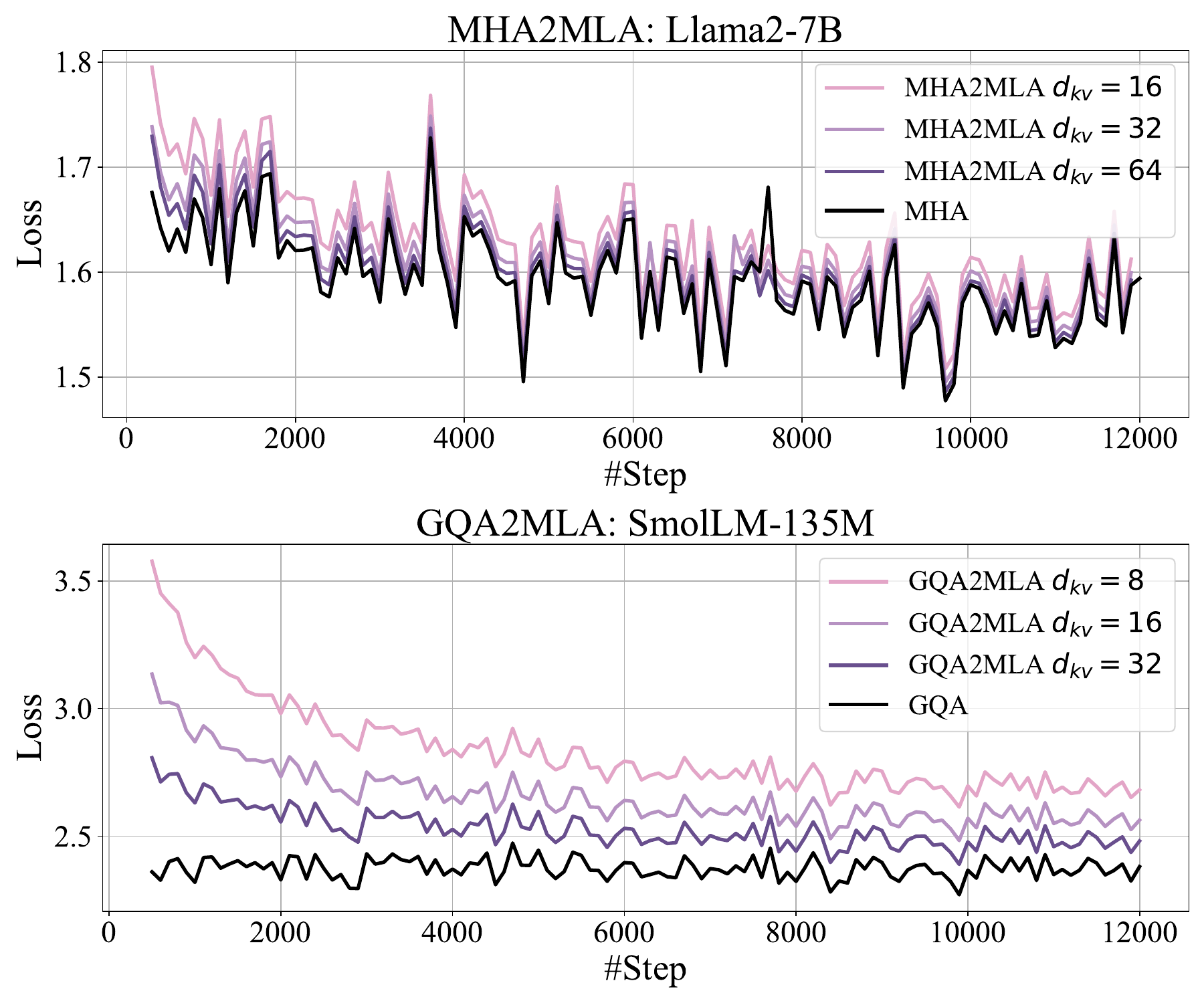}
  \caption{The fine-tuning loss curves under different KV cache storage ratios (with colors ranging from light to dark representing 12.5\%, 18.75\%, 31.25\%, and 100\%).}
  \label{fig:rank_loss}
  \vspace{-0.3cm}
\end{figure}

Next, as \(d_{kv}\) decreases (e.g., from 32 to 16 to 8), the KV cache reduction increases (i.e., from -68.75\% to -81.25\% to -87.5\%), but the performance loss becomes more challenging to recover through fine-tuning. 
\Cref{fig:rank_loss} shows the fine-tuning loss curves of 135M (representing GQA) and 7B (representing MHA) under different compression ratios. As the compression ratio increases, the loss difference from the baseline becomes larger. Additionally, we observe that the fluctuation trends of the loss curves are \textit{almost identical}, indicating that our architecture migration does not significantly harm the model's internal knowledge.

We also find that larger models experience less performance degradation when transitioning to the MLA architecture. For example, with compression down to 18.75\%, the performance drops by 2.24\% for 135M, 2.63\% for 360M, 1.43\% for 1B7, 0.30\% for 7B and 0.23\% for 13B, revealing the \textbf{potential scaling law of MHA2MLA}.
Finally, from the 135M model to the 13B model, the number of tokens required for fine-tuning is only about 0.6\% to 1\% of the pretraining tokens, demonstrating the data efficiency of our method.

Overall, whether using GQA2MLA or MHA2MLA, the architecture transition is achieved with minimal cost, resulting in efficient and economical inference.



\subsection{Long Context Tasks}
\begin{table}[t]
\centering
\begin{tabular}{l@{\hskip 5pt}p{2cm}@{\hskip 5pt}c@{\hskip 5pt}c}
  \toprule
  {\textbf{Model}} & \textbf{Precision} & \textbf{KV Mem.} & \textbf{Avg@LB} \\
  \midrule
  \rowcolor{gray!10}7B$_{\text{Llama2}}$
  & \raggedright BF16 & 100.0\% & 27.4 \\
  & \raggedright Int4$_{\text{HQQ}}$ & \multirow{2}{*}{-75.00\%} 
  & 27.5 \\
  & \raggedright Int4$_{\text{Quanto}}$ & & 27.3 \\
    \arrayrulecolor{gray!20}
  \hline

  & \raggedright Int2$_{\text{HQQ}}$ & \multirow{2}{*}{-87.50\%} & 21.2 \\
  & \raggedright Int2$_{\text{Quanto}}$ &  & 18.5 \\
  \arrayrulecolor{black}
  \hline
  \multirow{3}{*}{~$d_{kv}\!=\!64$} 
  & \raggedright BF16 & -68.75\% & 26.7 \\
  \arrayrulecolor{gray!20}
  \hline
  & \raggedright Int4$_{\text{HQQ}}$ & \multirow{2}{*}{-92.19\%}  & \bf 26.4 \\
  & \raggedright Int4$_{\text{Quanto}}$ & & \bf 26.3 \\
  \arrayrulecolor{black}
  \hline
  \multirow{3}{*}{~$d_{kv}\!=\!32$} 
  & \raggedright BF16 & -81.25\% & 26.0 \\
  \arrayrulecolor{gray!20}
  \hline
  & \raggedright Int4$_{\text{HQQ}}$ & \multirow{2}{*}{-95.31\%}  &\bf  25.8 \\
  & \raggedright Int4$_{\text{Quanto}}$ &  & \bf 25.5 \\
  \arrayrulecolor{black}
  \hline
  \multirow{3}{*}{~$d_{kv}\!=\!16$} 
  & \raggedright BF16 & -87.50\% & \bf 25.1 \\
  \arrayrulecolor{gray!20}
  \hline
  & \raggedright Int4$_{\text{HQQ}}$ & \multirow{2}{*}{-96.87\%}   & \bf 25.0 \\
  & \raggedright Int4$_{\text{Quanto}}$ &  & \bf 24.6 \\
  \arrayrulecolor{black}
    \bottomrule
\end{tabular}
\caption{Evaluation results of Llama2-7B and MHA2MLA on LongBench. \textbf{Bold} indicates compression ratios greater than or equal to Int2 quantization while also achieving performance higher than Int2.}
\vspace{-0.3cm}
\label{tab:long_bench}
\end{table}
\label{ssec:longbench_res}

\paragraph{Settings}
To evaluate the generative capabilities of the model, we adopt LongBench~\cite{acl/BaiLZL0HDLZHDTL24} as the benchmark for generation performance. All models are tested using a greedy decoding strategy. The context window size is determined based on the sequence length used during model fine-tuning. We use HQQ \cite{badri2023hqq} and Quanto\footnote{\url{https://huggingface.co/blog/quanto-introduction}} to set caches with different levels of precision to evaluate the performance of the original model as the baseline. Since our method is compatible with KV cache quantization, we also conduct additional experiments to assess the combined effect of both approaches.

\paragraph{Main Results}
As evidenced in \Cref{tab:long_bench}, MHA2MLA achieves competitive or superior efficiency-accuracy profiles compared to post-training quantization methods on LongBench. While 4-bit quantization incurs modest degradation (-0.2\% to -0.4\%) at comparable compression ratios, aggressive 2-bit quantization suffers severe performance collapse (-6.2\% to -9\%) despite 87.5\% KV cache reduction. In contrast, MHA2MLA alone attains 87.5\% compression (at \( d_{kv}\!=\!16 \)) with only 2.3\% accuracy loss, and further synergizes with 4-bit quantization to reach 92.19\%/96.87\% compression (\( d_{kv}\!=\!64/16 \)+Int4$_\text{HQQ}$) while limiting degradation to -1.0\%/-2.4\%, outperforming all 2-bit baselines. This highlights that MHA2MLA's latent space design remains orthogonal to numerical precision reduction, enabling \textbf{compound efficiency gains} without destructive interference.

\subsection{Ablation Study}
\label{ssec:ablation_study}

\begin{table}[t]
\centering
\small
\begin{tabular}{lcr@{\hspace{2pt}}l}
  \toprule
  \textbf{Model} & \textbf{Tokens} & \multicolumn{2}{l}{\textbf{Avg@CS}} \\
  \midrule
  \rowcolor{gray!10}135M$_{\text{SmolLM}}$ & 600B & 44.50  \\
  \textit{- full-rope} & \multirow{5}{*}{6B}  & 44.42 \\
  - $\mathcal{S}_{\text{high}}$ &   & 43.60&\textsubscript{-0.82} \\
  - $\mathcal{S}_{\text{low}}$ &   & 39.17&\textsubscript{-5.25} \\
  - $\mathcal{S}_{\text{uniform}}$ &   &\bf 44.01&\textsubscript{-0.41} \\
  - $\mathcal{S}_{\text{2-norm}}$ &   &  43.73&\textsubscript{-0.69} \\
  \arrayrulecolor{gray!20}
  \hline
  - $\mathcal{S}_{\text{high}}$ + SVD\textsubscript{joint} &  \multirow{4}{*}{6B} & 40.85&\textsubscript{-3.57} \\
  - $\mathcal{S}_{\text{uniform}}$ + SVD\textsubscript{joint} &   & 41.79&\textsubscript{-2.63} \\
  - $\mathcal{S}_{\text{2-norm}}$ + SVD\textsubscript{joint} &   & \bf 42.18&\textsubscript{-2.24} \\
  - $\mathcal{S}_{\text{2-norm}}$ + SVD\textsubscript{split} & & 41.27 & \textsubscript{-3.15} \\
  \arrayrulecolor{black}
  \midrule
  \rowcolor{gray!10}1B7$_{\text{SmolLM}}$ & 1T & 55.90 \\
  \textit{- full-rope} & \multirow{5}{*}{6B}  & 55.71 \\
  - $\mathcal{S}_{\text{high}}$ &   & 54.80&\textsubscript{-0.91} \\
  - $\mathcal{S}_{\text{low}}$ &   & 53.84&\textsubscript{-1.87} \\
  - $\mathcal{S}_{\text{uniform}}$ &   &\bf  55.30&\textsubscript{-0.41} \\
  - $\mathcal{S}_{\text{2-norm}}$ &   & 54.98&\textsubscript{-0.73} \\
  \arrayrulecolor{gray!20}
  \hline
  - $\mathcal{S}_{\text{high}}$ + SVD\textsubscript{joint} & \multirow{4}{*}{6B}  & 54.17&\textsubscript{-1.54} \\
  - $\mathcal{S}_{\text{uniform}}$ + SVD\textsubscript{joint} &   & 54.27&\textsubscript{-1.44} \\
  - $\mathcal{S}_{\text{2-norm}}$ + SVD\textsubscript{joint} &   & \bf 54.28&\textsubscript{-1.43} \\
  - $\mathcal{S}_{\text{2-norm}}$ + SVD\textsubscript{split} &   & 52.90  & \textsubscript{-2.81} \\
  \arrayrulecolor{black}
  \bottomrule
\end{tabular}
\caption{Reasoning ability of ablation studies. The results of other models are provided in Appendix~\ref{app:other_lb}.}
\vspace{-0.4cm}
\label{tab:partial_rope}
\end{table}
\paragraph{Four Partial-RoPE strategies: $\mathcal{S}_{\text{high}}$, $\mathcal{S}_{\text{low}}$, $\mathcal{S}_{\text{uniform}}$, $\mathcal{S}_{\text{2-norm}}$}
\Cref{tab:partial_rope} presents the results of four strategies for converting full-RoPE to partial-RoPE. 
First, when comparing the four strategies with full-RoPE, we observed that the low-frequency retention strategy, \(\mathcal{S}_{\text{low}}\), incurred the greatest performance loss (a reduction of -5.25\%@135M and -1.87\%@1B7), whereas the high-frequency retention strategy, \(\mathcal{S}_{\text{high}}\), experienced significantly less degradation (a reduction of -0.82\%@135M and -0.91\%@1B7), underscoring the importance of high-frequency subspaces. Both \(\mathcal{S}_{\text{uniform}}\) and \(\mathcal{S}_{2\text{-norm}}\) yielded better performance, the \(\mathcal{S}_{\text{uniform}}\) preserves subspaces across the frequency spectrum, while the \(\mathcal{S}_{2\text{-norm}}\) retains subspaces based on their contribution to the attention scores. 
We choose \( \mathcal{S}_{2\text{-norm}} \) as the default configuration because the removed subspaces (i.e., NoPE) are more suitable for the (SVD-based) low-rank approximation.

\paragraph{Two SVD-based low-rank approximations: SVD\textsubscript{split}, SVD\textsubscript{joint}}
The last two rows of each group in \Cref{tab:partial_rope} compare the effects of the two SVD methods. 
We observe that, on both LLMs, the SVD\(_{\text{joint}}\) method consistently outperforms SVD\(_{\text{split}}\), yielding an average performance improvement of 0.91\% on the 135M model and 1.38\% on the 1B7 model. 
It indicates that SVD\(_{\text{joint}}\) emerges as the clear default choice.

\section{Related Work}
\label{sec:related-work}

\paragraph{Efficient Attention Architectures}  
The standard Multi-Head Attention (MHA, \citeyear{nips/VaswaniSPUJGKP17}) mechanism's quadratic complexity in context length has spurred numerous efficiency innovations. 
While MHA remains foundational, variants like Multi-Query Attention (MQA) and Grouped-Query Attention (GQA, \citeyear{emnlp/AinslieLJZLS23}) reduce memory overhead by sharing keys/values across heads—albeit at the cost of parameter pruning and performance degradation. 
Parallel efforts, such as Linear Transformers \cite{naacl/GuoQLSXZ19,icml/KatharopoulosV020, iclr/ChoromanskiLDSG21}, RWKV \cite{emnlp/PengAAAABCCCDDG23}, and Mamba \cite{corr/abs-2312-00752}, replace softmax attention with linear recurrences or state-space models, but struggle to match the expressiveness of standard attention in autoregressive generation.  

Multi-Head Latent Attention (MLA, \citeyear{corr/abs-2405-04434})  distinguishes itself by compressing KV caches into low-rank latent vectors without pruning attention parameters. 
Our work bridges MLA with mainstream architectures (MHA/GQA), enabling seamless migration via data-efficient fine-tuning.  
Notably, while many linear attention variants abandon softmax query-key interactions (e.g., through kernel approximations), architectures preserving a query-key dot product structure—even in factorized forms—remain compatible with our MHA2MLA framework.

\paragraph{Economical Key-Value Cache}  
The memory footprint of KV caches has become a critical bottleneck for long-context inference. 
Recent advances fall into three categories:  

\textit{Innovative Architecture} methods like MLA~\cite{corr/abs-2405-04434}, MiniCache~\cite{nips/LiuLPHHZ24}, and MLKV~\cite{corr/abs-2406-09297} share or compress KV representations across layers or heads. 
While effective, cross-layer sharing risks conflating distinct attention patterns, potentially harming task-specific performance. 
Only MLA has been successfully validated in Deepseek's LLMs.

\textit{Quantization} techniques such as GPTQ~\cite{corr/abs-2210-17323}, FlexGen~\cite{icml/0007ZYLRCLRSZ23}, and KIVI~\cite{icml/LiuYJZXBC024} store KV caches in low-bit formats (e.g., 2-bit), achieving memory savings with precision loss.  

\textit{Dynamic Pruning} approaches like A2SF~\cite{corr/abs-2407-20485} and SnapKV~\cite{nips/LiHYVLYCLC24} prune ``less important'' tokens from the KV cache. However, token pruning risks discarding critical long-range dependencies, while head pruning (e.g., SliceGPT~\cite{iclr/AshkboosCNHH24}, Sheared~\cite{conf/iclr/XiaGZ024}, and Simple Pruning~\cite{conf/iclr/Sun0BK24}) irreversibly reduces model capacity.  

Our MHA2MLA method achieves the migration of standard Transformer-based LLMs to the more economical MLA architecture and has demonstrated its ability to integrate with KV quantization techniques to realize a \textasciitilde 97\% cache saving. It is also theoretically compatible with other methods like pruning.





\section{Conclusion}  
This work addresses the critical challenge of adapting pre-trained MHA-based LLMs (or variants) to the KV-cache-efficient MLA architecture. By introducing MHA2MLA with contribution-aware partial-RoPE removal and SVD-driven low-rank projection, we achieve near-lossless compression of KV cache (up to 96.87\% size reduction for Llama2-7B) while requiring only 0.6\%~to~1\% of training data. The framework demonstrates strong compatibility with existing compression techniques and maintains commonsense reasoning and long-context processing capabilities, offering a practical pathway for deploying resource-efficient LLMs without sacrificing performance. Our results underscore the feasibility of architectural migration for LLMs through targeted parameter reuse and data-efficient fine-tuning.

\section*{Limitations}  

\paragraph{Verification on More LLMs}  
Considering that MHA2MLA can significantly reduce inference costs, it is worthwhile to validate it on larger and more diverse open-source LLMs. However, constrained by our computation resources, models like Llama3 require fine-tuning on a 128K context length to mitigate performance degradation from continued training, so we did not perform such experiments. Furthermore, since Deepseek has not yet open-sourced the tensor-parallel inference framework for MLA, it is currently challenging to explore models larger than 7B. This will be addressed in our future work.

\paragraph{Parameter-Efficient MHA2MLA Fine-tuning}  
This paper primarily focuses on the data efficiency of MHA2MLA. Since the architectural transformation does not involve the Feed-Forward (FFN) module, future work could explore parameter-efficient MHA2MLA fine-tuning, for example by freezing the FFN module and/or freezing the parameters in the queries and keys that correspond to the retained RoPE. This could further reduce the cost of the MHA2MLA transition.

\section*{Acknowledgments}
The authors wish to thank all reviewers for their
helpful comments and suggestions.
The corresponding authors are Yuanbin Wu, Xipeng Qiu, Qi Zhang, Tao Gui. 
This work was partially funded by Guangdong S\&T Program 2024B0101050003, National Natural Science Foundation of China (No.62076069,62206057,61976056), Shanghai Rising-Star Program (23QA1400200), and Natural Science Foundation of Shanghai (23ZR1403500), Program of Shanghai Academic Research Leader under grant 22XD1401100.
The computations in this research were
performed using the CFFF platform of Fudan University.

\bibliography{main}

\appendix

\clearpage
\section{The Calculation of 2-norm Score}
\label{app:2_norm}


To compute the 2-norm scores for each attention head, we selected 1,024 samples from the training dataset. The proportions of the subsets and sequence length used during the 2-norm computation are consistent with those used during fine-tuning. First, we calculate the query vectors and key vectors for each head. Then, for each rotational subspace of the vectors, we compute the 2-norm scores. Finally, the 2-norm scores of the query and key vectors are aggregated within each subspace. If the model employs Grouped-Query Attention (GQA), the 2-norm scores are averaged within each GQA group, and the scores are shared between the groups.

\section{Inference Process of MHA2MLA}
\label{app:mha2mla_infer}
During inference in the MHA2MLA model, our input includes the hidden representation \( x_i \) of the \( i \)-th token, as well as the previously stored \(\bm{k}_{<i, \text{rope}}^{(h)}\) and \(\bm{c}_{<i, \text{kv}}\) in the KV cache for the first \( i-1 \) tokens.  

During the inference, our goal is to compute the \( h \)-th head's dot product of these two parts $\bm{q}_{i,\text{rope}}^{(h)} \bm{k}_{\le i,\text{rope}}^{(h)\top}$ and $\bm{q}_{i,\text{nope}}^{(h)} \bm{k}_{\le i,\text{nope}}^{(h)\top}$.
For the RoPE part, we can easily extract \( \bm{W}_{q, \text{rope}}^{(h)} \) and \( \bm{W}_{k, \text{rope}}^{(h)} \) from the pre-trained parameter matrices \( \bm{W}_q^{(h)} \) and \( \bm{W}_k^{(h)} \) (i.e., the rows corresponding to the subspace that retains RoPE) and then obtain the result through a linear transformation:
\begin{align*}  
\bm{q}_{i,\text{rope}}^{(h)} &= \bm{x}_i\bm{W}_{q,
\text{rope}}^{(h)}\\
\bm{k}_{i,\text{rope}}^{(h)} &= \bm{x}_i\bm{W}_{k,
\text{rope}}^{(h)}\\
\bm{k}_{\le i,\text{rope}}^{(h)} &= [\bm{k}_{<i, \text{rope}}^{(h)}, ~\bm{k}_{i,\text{rope}}^{(h)}]
\\ \to ~&\bm{q}_{i,\text{rope}}^{(h)} \bm{k}_{\le i,\text{rope}}^{(h)\top}.
\end{align*}
Note that \(\bm{k}_{<i, \text{rope}}^{(h)}\) is already stored in the KV cache and can be directly retrieved.

For the NoPE part, \(\bm{q}_{i,\text{nope}}^{(h)}\) can still be easily obtained through a linear transformation $\bm{W}_{q,\text{nope}}^{(h)}$ which extracted from the pre-trained parameter matrix \( \bm{W}_q^{(h)} \) by separating the rows corresponding to the subspace with RoPE removed.  
However, \(\bm{k}_{i,\text{nope}}^{(h)}\) requires two linear transformations:  a \textit{dimensionality reduction} transformation using \(\bm{W}_{dkv}\), and a \textit{dimensionality expansion} transformation using \(\bm{W}_{uk}^{(h)}\).  
Note that \(\bm{W}_{dkv}\) is shared across all heads in the current layer, and both \(\bm{W}_{dkv}\) and \(\bm{W}_{uk}^{(h)}\) are constrained by the SVD decomposition of the pre-trained parameter matrices \(\bm{W}_{k,\text{nope}}^{(h)}\) and \(\bm{W}_{v}^{(h)}\), preserving most of the pre-trained knowledge:
\begin{align*}  
\bm{q}_{i,\text{nope}}^{(h)} &= \bm{x}_i\bm{W}_{q,
\text{nope}}^{(h)}\\
\bm{c}_{i, kv} &= \bm{x}_i\bm{W}_{dkv,
}\\
\bm{k}_{i,\text{nope}}^{(h)} &= \bm{c}_{i, kv}\bm{W}_{uk}^{(h)}\\
\bm{k}_{<i, \text{nope}}^{(h)} &= \bm{c}_{<i, kv}\bm{W}_{uk}^{(h)}.
\end{align*}
During inference, the NoPE part can also leverage the standard MLA matrix merging algorithm to reduce memory consumption:
\begin{align*}
\bm{k}_{\le i, \text{nope}}^{(h)} &= [\bm{c}_{<i, kv},~ \bm{c}_{i, kv}]\bm{W}_{uk}^{(h)}\\
 \bm{q}_{i,\text{nope}}^{(h)} \bm{k}_{\le i,\text{nope}}^{(h)\top} & = (\bm{x}_i\bm{W}_{q,
\text{nope}}^{(h)})  (\bm{c}_{\le i, kv}\bm{W}_{uk}^{(h)})^\top \\
 & = \bm{x}_i (\bm{W}_{q,
\text{nope}}^{(h)} \bm{W}_{uk}^{(h)\top}) \bm{c}_{\le i, kv}^\top.
\end{align*}
We can pre-multiply the parameter matrices $(\bm{W}_{q,
\text{nope}}^{(h)} \bm{W}_{uk}^{(h)\top})$, and let $\bm{c}_{ i, q}^{(h)} = \bm{x}_i (\bm{W}_{q, \text{nope}}^{(h)} \bm{W}_{uk}^{(h)\top})$.
In the end, the output of MHA2MLA is as follows:
\begin{align*}    
& \bm{v}_i^{(h)} = \bm{c}_{i, kv}\bm{W}_{uv}^{(h)}\\
&\bm{o}_i^{(h)}\!=\!\text{Softmax}\!\left(\bm{q}_{i,\text{rope}}^{(h)}\bm{k}_{\le i,\text{rope}}^{(h)\top}\!+\!\bm{c}_{i, q}^{(h)}\bm{c}_{\le i, kv}^\top\!\right) \bm{c}_{\le i, kv} \nonumber 
    \\ &\text{MHA2MLA}(\bm{x}_i) = \left[\dots, \bm{o}_i^{(h)}\bm{W}_{uv}^{(h)},  \dots\right]  \bm{W}_o.
\end{align*}
Where $\bm{W}_{uv}^{(h)}$ and $\bm{W}_o$ can also perform matrix merging to make inference more economical.

\paragraph{Why doesn't MHA2MLA perform low-rank representation on the query as DeepSeek does?}
Firstly, we found that the economical inference of MLA is not affected even if $\bm{W}_{q,\text{nope}}^{(h)}$ is not decomposed into a dimension-reducing matrix (e.g., $\bm{W}_{dq}$) and a dimension-increasing matrix (e.g., $\bm{W}_{uq}^{(h)}$). 
Secondly, decomposing $\bm{W}_{q,\text{nope}}^{(h)}$ introduces additional architectural migration loss (approximation loss) and further reduces the number of LLM parameters. 
Therefore, we believe there is no need to decompose $\bm{W}_{q,\text{nope}}^{(h)}$ within the MHA2MLA framework.

\section{The Details of Fine-tuning}
\label{app:ft_details}

\paragraph{Data}


We fine-tune our model using the pretraining corpus from SmolLM\footnote{\url{https://huggingface.co/blog/smollm}}. 
The dataset consists of fineweb-edu-dedup, cosmopedia-v2, python-edu, open-web-math, and StackOverflow. The first three datasets are part of the smollm-corpus\footnote{\url{https://huggingface.co/datasets/HuggingFaceTB/smollm-corpus}} curated by HuggingFaceTB. Fineweb-edu-dedup is a high-quality dataset filtered by HuggingFaceTB from education-related webpages. Similarly, HuggingFaceTB filtered Python code snippets from The Stack to construct the python-edu dataset. Cosmopedia-v2 is a high-quality dataset generated by a model based on 34,000 topics defined by BISAC book classifications. Additionally, open-web-math\footnote{\url{https://huggingface.co/datasets/open-web-math/open-web-math}} and StackOverflow\footnote{\url{https://huggingface.co/datasets/bigcode/stackoverflow-clean}} are sourced from high-quality mathematical texts available online and posts from StackOverflow, respectively.

\paragraph{Hyperparameters}
\begin{table*}[t]
\centering
\small
\setlength\tabcolsep{3pt}
\begin{tabular}{l@{}lccccc}
  \toprule
  \multicolumn{2}{l}{\textbf{Metrics}} & \textbf{135M$_{\text{SmolLM}}$} & \textbf{360M$_{\text{SmolLM}}$} & \textbf{1B7$_{\text{SmolLM}}$} & \textbf{7B$_{\text{Llama2}}$} & \textbf{13B$_{\text{Llama2}}$} \\
  \midrule
  \multicolumn{2}{l}{n\_batch $\times$ n\_gpu} & 64$\times$4 & 64$\times$4 & 32$\times$8 & 16$\times$16 & 8$\times$32 \\
  \multicolumn{2}{l}{Learning Rate} & 1e-4 & 1e-4 & 1e-4 & 1e-4 & 1e-4 \\
  \multicolumn{2}{l}{Hardware} & NVIDIA L20Y & NVIDIA L20Y & NVIDIA L20Y & NVIDIA L20Y & NVIDIA L20Y \\
  \multicolumn{2}{l}{Steps} & 12000 & 12000 & 12000 & 12000 & 12000 \\
  \multicolumn{2}{l}{Warmup ratio} & 10.0\% & 10.0\% & 10.0\% & 10.0\% & 10.0\% \\
  \multicolumn{2}{l}{Decay} & 16.7\% & 16.7\% & 16.7\% & 16.7\% & 16.7\% \\
  \multicolumn{2}{l}{Time} & 4h & 8h & 16h & 28h & 36h \\
  \multicolumn{2}{l}{Seqlen} & 2048 & 2048 & 2048 & 4096 & 4096 \\
  \multirow{4}{*}{\#Param.} & $d_{kv}\!=\!128/256^\dag$ & 134.52M & 361.82M & 1.71B & 6.61B$^\dag$& 13.02B$^\dag$ \\
  & $d_{kv}\!=\!32/64^\dag$  & 130.99M & 351.38M & 1.67B & 6.37B$^\dag$ & 12.56B$^\dag$ \\
  & $d_{kv}\!=\!16/32^\dag$  & 129.64M & 347.38M & 1.59B & 5.99B$^\dag$ & 11.80B$^\dag$ \\
  & $d_{kv}\!=\!8/16^\dag$   & 128.97M & 345.39M & 1.56B & 5.79B$^\dag$ & 11.43B$^\dag$ \\
  \arrayrulecolor{black}
  \bottomrule
\end{tabular}
\caption{Training detail information across different models.}
\label{tab:Hyperparameters}
\end{table*}




The fine-tuning hyperparameters for models of all sizes are listed in \Cref{tab:Hyperparameters}. The training process employs a warmup phase followed by a decay strategy. A 1-sqrt decay strategy is applied to ensure a smooth and gradual reduction.

\begin{table*}[t]
\centering
\small
\begin{tabular}{l@{}lr@{\hspace{2pt}}lcccccc}
  \toprule
  \multicolumn{2}{l}{\textbf{Model}}  & \multicolumn{2}{c}{\textbf{Avg.}} & \textbf{MMLU} & \textbf{ARC} & \textbf{PIQA} & \textbf{HS} & \textbf{OBQA} & \textbf{WG}\\
  \midrule
  \rowcolor{gray!10}135M & $r$=32 & \multicolumn{2}{l}{44.42}  & 29.91 & 41.71 & 68.28 & 41.33 & 33.80 & 51.46 \\
  \arrayrulecolor{gray!20}
  \hline
  \arrayrulecolor{gray!20}
  \hline
  \multirow{4}{*}{- $\mathcal{S}_{\text{high}}$} 
  & $r$=1 & 42.88 &\textsubscript{-1.54}& 29.24 & 40.15 & 66.81 & 37.90 & 33.40 & 49.80 \\
  & $r$=2 & 43.07 &\textsubscript{-1.35}& 29.73 & 40.60 & 67.25 & 38.82 & 32.40 & 49.64 \\
  & $r$=4 & 43.60 &\textsubscript{-0.82}& 29.87 & 41.29 & 67.08 & 39.58 & 32.80 & 50.99 \\
  & $r$=8 & 43.90 &\textsubscript{-0.52}& 29.79 & 40.89 & 68.01 & 40.71 & 33.40 & 50.59 \\
  \arrayrulecolor{gray!20}
  \hline
  \multirow{4}{*}{- $\mathcal{S}_{\text{low}}$} 
  & $r$=1 & 39.85 &\textsubscript{-4.57}& 27.72 & 36.56 & 62.95 & 33.88 & 28.20 & 49.80 \\
  & $r$=2 & 39.72 &\textsubscript{-4.70}& 27.36 & 36.86 & 63.76 & 33.85 & 27.80 & 48.70 \\
  & $r$=4 & 39.17 &\textsubscript{-5.25}& 27.67 & 35.33 & 62.30 & 33.32 & 27.60 & 48.78 \\
  & $r$=8 & 42.36 &\textsubscript{-2.06}& 29.33 & 39.37 & 66.70 & 38.13 & 31.00 & 49.64 \\
  \arrayrulecolor{gray!20}
  \hline
  \multirow{4}{*}{- $\mathcal{S}_{\text{uniform}}$} 
  & $r$=1 & 42.72 &\textsubscript{-1.70}& 29.34 & 40.20 & 66.76 & 37.60 & 32.60 & 49.80 \\
  & $r$=2 & 43.50 &\textsubscript{-0.92}& 29.41 & 41.30 & 67.63 & 39.31 & 33.40 & 49.96 \\
  & $r$=4 & 44.01 &\textsubscript{-0.41}& 29.79 & 41.09 & 67.95 & 40.54 & 34.20 & 50.51 \\
  & $r$=8 & 43.79 &\textsubscript{-0.66}& 29.85 & 40.72 & 67.57 & 40.84 & 32.80 & 50.99 \\
  \arrayrulecolor{gray!20}
  \hline
  \multirow{4}{*}{- $\mathcal{S}_{\text{2-norm}}$} 
  & $r$=1 & 43.27 &\textsubscript{-1.15}& 29.58 & 40.83 & 67.25 & 39.14 & 33.00 & 49.80 \\
  & $r$=2 & 43.77 &\textsubscript{-0.65}& 29.82 & 40.76 & 68.28 & 39.32 & 34.40 & 50.04 \\
  & $r$=4 & 43.73 &\textsubscript{-0.69}& 30.00 & 41.29 & 68.17 & 39.83 & 33.20 & 49.88 \\
  & $r$=8 & 44.18 &\textsubscript{-0.24}& 30.01 & 41.52 & 68.12 & 40.70 & 34.00 & 50.75 \\
  \arrayrulecolor{black}
  \bottomrule
\end{tabular}
\caption{The impact of positional encoding dimensionality on model performance.}
\label{tab:pe_dim}
\end{table*}

\section{Ablation Study on Partial-RoPE Dimensions}
\label{app:pe-dim}

To better determine the strategy and dimensionality for partial-RoPE, we conducted an ablation study on the number of RoPE dimensions using the 135M$_{\text{SmolLM}}$ model. The experimental results are presented in \Cref{tab:pe_dim}. By comparing the performance of four different strategies in varying dimensionalities, we observed that the low-frequency strategy, $\mathcal{S}_{\text{low}}$, suffered significant performance degradation (-11.8\%) when the dimensionality was relatively low ($\leq 4$). In contrast, both $\mathcal{S}_{\text{uniform}}$ and $\mathcal{S}_{\text{2-norm}}$ consistently demonstrated superior performance regardless of dimensionality. Furthermore, increasing the dimensionality from 4 to 8 provided negligible performance gains. Based on these findings, we selected a dimensionality of 4 for partial-RoPE.

\section{Detailed Results}
\label{app:other_lb}
\begin{table*}[t]
\centering
\small
\setlength\tabcolsep{3pt}
\begin{tabular}{llrrrrrrrrrrrrrrrrrr}
  \toprule
  \multirow{2}{*}{\textbf{$d_{kv}$}} & \multirow{2}{*}{\textbf{Precision}} & \multirow{2}{*}{\textbf{KV}} & \multirow{2}{*}{\textbf{Avg.}} & \multicolumn{3}{c}{\textbf{S-Doc QA}} & \multicolumn{3}{c}{\textbf{M-Doc QA}} & \multicolumn{3}{c}{\textbf{Summ.}} & \multicolumn{3}{c}{\textbf{Few-shot}} & \multicolumn{2}{c}{\textbf{Synth.}} & \multicolumn{2}{c}{\textbf{Code}} \\
  \cmidrule(lr){5-7} \cmidrule(lr){8-10} \cmidrule(lr){11-13} \cmidrule(lr){14-16} \cmidrule(lr){17-18} \cmidrule(lr){19-20} 
  & & & & \textbf{A} & \textbf{B} & \textbf{C} & \textbf{D} & \textbf{E} & \textbf{F} & \textbf{G} & \textbf{H} & \textbf{I} & \textbf{J} & \textbf{K} & \textbf{L} & \textbf{M} & \textbf{N} & \textbf{O} & \textbf{P} \\ 
  \midrule
  \rowcolor{gray!10} \multicolumn{20}{c}{\textit{\textbf{7B$_{\text{Llama2}}$ (Length=4K)}}} \\
  & \raggedright BF16 & 100.0\% & 27.4 & 15.1 & 9.6 & 21.1 & 7.5 & 9.7 & 3.7 & 26.7 & 20.5 & 3.2 & 65.5 & 87.5 & 34.1 & 1.9 & 6.6 & 66.5 & 59.4 \\
  & \raggedright Int4$_{\text{HQQ}}$ & \multirow{2}{*}{-75.00\%} 
  & 27.5 & 16.1 & 9.1 & 22.0 & 7.3 & 9.9 & 3.6 & 26.5 & 21.1 & 3.4 & 65.5 & 87.2 & 34.3 & 1.5 & 6.7 & 66.0 & 59.9 \\
  & \raggedright Int4$_{\text{Quanto}}$ & & 27.3 & 14.4 & 9.5 & 20.5 & 7.5 & 9.7 & 3.5 & 25.8 & 20.7 & 3.1 & 65.5 & 87.7 & 34.3 & 1.4 & 7.3 & 66.8 & 59.3 \\
    \arrayrulecolor{gray!20}
  \hline
  & \raggedright Int2$_{\text{HQQ}}$ & \multirow{2}{*}{-87.50\%} & 21.2 & 18.0 & 5.5 & 12.6 & 7.5 & 8.4 & 3.2 & 12.6 & 18.6 & 0.9 & 56.5 & 73.3 & 27.0 & 1.8 & 6.1 & 34.5 & 52.9 \\
  & \raggedright Int2$_{\text{Quanto}}$ &  & 18.5 & 9.4 & 6.2 & 12.7 & 6.8 & 6.7 & 3.3 & 5.9 & 17.2 & 0.4 & 61.0 & 63.9 & 26.0 & 1.4 & 2.7 & 42.4 & 30.5 \\
  \arrayrulecolor{black}
  \hline
  \multirow{3}{*}{$64$} 
  & \raggedright BF16 & -68.75\% & 26.7 & 12.2 & 9.4 & 22.5 & 7.5 & 11.7 & 4.2 & 26.5 & 18.9 & 20.2 & 58.0 & 83.6 & 35.0 & 1.7 & 5.5 & 57.1 & 52.8 \\
  \arrayrulecolor{gray!20}
  \hline
  & \raggedright Int4$_{\text{HQQ}}$ & \multirow{2}{*}{-92.19\%}  & \bf 26.4 & 12.9 & 9.3 & 22.7 & 7.8 & 12.1 & 3.3 & 26.5 & 18.7 & 18.6 & 58.0 & 82.4 & 35.5 & 1.4 & 4.9 & 56.2 & 51.8 \\
  & \raggedright Int4$_{\text{Quanto}}$ & & \bf 26.3 & 9.5 & 8.7 & 22.7 & 7.6 & 11.0 & 4.0 & 26.0 & 18.3 & 19.8 & 58.5 & 84.4 & 35.3 & 1.3 & 5.1 & 56.7 & 51.4\\
  \arrayrulecolor{black}
  \hline
  \multirow{3}{*}{$32$} 
  & \raggedright BF16 & -81.25\% & 26.0 & 13.4 & 8.7 & 21.2 & 5.9 & 9.9 & 2.5 & 25.3 & 19.2 & 17.6 & 65.5 & 85.5 & 25.5 & 3.0 & 7.0 & 54.0 & 51.4\\
  \arrayrulecolor{gray!20}
  \hline
  & \raggedright Int4$_{\text{HQQ}}$ & \multirow{2}{*}{-95.31\%}  &\bf 25.8 & 13.6 & 9.1 & 20.6 & 6.0 & 10.2 & 2.5 & 25.0 & 18.4 & 16.4 & 65.5 & 85.1 & 25.4 & 3.1 & 6.9 & 53.4 & 51.0 \\
  & \raggedright Int4$_{\text{Quanto}}$ &  & \bf 25.5 & 13.4 & 8.0 & 21.2 & 6.4 & 10.1 & 3.0 & 24.3 & 17.1 & 17.1 & 65.0 & 85.1 & 26.1 & 3.6 & 6.2 & 52.7 & 49.0 \\
  \arrayrulecolor{black}
  \hline
  \multirow{3}{*}{$16$} 
  & \raggedright BF16 & -87.50\% & \bf 25.1 & 13.2 & 8.7 & 21.6 & 7.1 & 9.1 & 3.7 & 24.1 & 18.4 & 20.3 & 57.5 & 86.0 & 33.3 & 0.1 & 9.0 & 43.9 & 44.9 \\
  \arrayrulecolor{gray!20}
  \hline
  & \raggedright Int4$_{\text{HQQ}}$ & \multirow{2}{*}{-96.87\%}   & \bf 25.0 & 13.7 & 8.8 & 23.7 & 7.1 & 9.2 & 4.5 & 22.8 & 18.7 & 18.4 & 57.5 & 86.6 & 32.1 & 0.1 & 8.8 & 43.5 & 44.3 \\
  & \raggedright Int4$_{\text{Quanto}}$ &  & \bf 24.6 & 9.9 & 8.4 & 22.3 & 7.2 & 9.0 & 4.2 & 22.6 & 18.4 & 18.6 & 57.0 & 85.4 & 33.6 & 0.4 & 8.8 & 43.5 & 45.0 \\
  \arrayrulecolor{black}
  
  \rowcolor{gray!10} \multicolumn{20}{c}{\textit{\textbf{1B7$_{\text{SmolLM}}$ (Length=2K)}}} \\
  & \raggedright BF16 & 100.0\% & 18.7 & 2.6 & 6.3 & 19.9 & 5.4 & 8.6 & 2.7 & 23.5 & 18.4 & 20.2 & 46.5 & 70.2 & 32.4 & 2.2 & 3.2 & 21.3 & 16.5 \\
  & \raggedright Int4$_{\text{HQQ}}$ & \multirow{2}{*}{-75.00\%} 
  & 18.6 & 2.5 & 6.2 & 19.1 & 5.5 & 8.2 & 2.7 & 23.4 & 18.3 & 20.0 & 46.5 & 69.4 & 32.1 & 2.7 & 3.2 & 21.5 & 16.0 \\
  & \raggedright Int4$_{\text{Quanto}}$ & & 18.6 & 2.6 & 6.2 & 17.4 & 5.1 & 8.6 & 2.6 & 23.0 & 18.1 & 20.1 & 46.0 & 70.2 & 31.9 & 2.9 & 3.6 & 21.9 & 16.7 \\
  & \raggedright Int2$_{\text{HQQ}}$ & \multirow{2}{*}{-87.50\%} & 16.3 & 2.5 & 5.6 & 13.0 & 4.8 & 7.5 & 2.7 & 14.8 & 16.3 & 9.3 & 46.0 & 70.4 & 26.9 & 2.6 & 3.4 & 18.3 & 16.8 \\
  & \raggedright Int2$_{\text{Quanto}}$ &  & 13.3 & 1.6 & 3.8 & 10.3 & 3.9 & 7.3 & 1.4 & 5.9 & 13.4 & 6.3 & 40.0 & 64.3 & 14.6 & 3.1 & 3.5 & 15.6 & 17.5 \\
  \hline
  \multirow{3}{*}{$32$} 
  & \raggedright BF16 & -68.75\% & 16.3 & 2.6 & 6.1 & 17.8 & 5.3 & 9.2 & 2.4 & 21.8 & 14.7 & 18.8 & 52.5 & 55.6 & 28.8 & 1.7 & 4.7 & 12.5 & 5.7 \\
  & \raggedright Int4$_{\text{HQQ}}$ & \multirow{2}{*}{-92.19\%} & 15.7 & 2.3 & 6.4 & 16.4 & 5.2 & 8.8 & 2.3 & 20.5 & 14.2 & 16.6 & 52.5 & 56.7 & 27.8 & 1.5 & 4.2 & 11.9 & 4.4\\
  & \raggedright Int4$_{\text{Quanto}}$ & & 15.7 & 2.3 & 6.0 & 16.5 & 5.3 & 8.8 & 2.1 & 22.1 & 14.5 & 17.5 & 50.5 & 55.0 & 27.6 & 1.8 & 3.2 & 13.3 & 5.5 \\
  \hline
  \multirow{3}{*}{$16$} 
  & \raggedright BF16 & -81.25\% & 15.5 & 2.4 & 6.2 & 17.1 & 5.5 & 9.2 & 2.5 & 21.0 & 15.2 & 16.5 & 47.5 & 53.9 & 31.4 & 1.3 & 3.3 & 9.5 & 5.3 \\
  & \raggedright Int4$_{\text{HQQ}}$ & \multirow{2}{*}{-95.31\%} & 15.3 & 2.4 & 5.7 & 17.0 & 4.8 & 9.0 & 2.1 & 20.0 & 15.5 & 16.8 & 47.5 & 53.1 & 30.1 & 2.0 & 3.4 & 10.6 & 5.3 \\
  & \raggedright Int4$_{\text{Quanto}}$ &  & 15.1 & 2.3 & 5.9 & 16.0 & 6.0 & 9.4 & 2.5 & 19.1 & 14.4 & 15.5 & 47.5 & 52.5 & 28.4 & 2.0 & 3.2 & 11.2 & 5.3 \\
  \hline
  \multirow{3}{*}{$8$} 
  & \raggedright BF16 & -87.50\% & 14.0 & 2.6 & 5.6 & 16.5 & 5.1 & 8.9 & 2.1 & 19.8 & 15.7 & 14.2 & 40.0 & 51.0 & 28.0 & 2.1 & 3.3 & 7.1 & 2.6\\
  & \raggedright Int4$_{\text{HQQ}}$ & \multirow{2}{*}{-96.87\%} & 13.8 & 2.6 & 5.0 & 15.4 & 4.5 & 9.5 & 2.5 & 20.5 & 14.8 & 14.0 & 40.0 & 48.2 & 27.1 & 1.8 & 4.2 & 7.6 & 3.1 \\
  & \raggedright Int4$_{\text{Quanto}}$ &  & 13.9 & 2.5 & 5.4 & 16.6 & 4.8 & 8.9 & 2.3 & 19.3 & 14.6 & 15.8 & 40.0 & 50.2 & 26.4 & 1.1 & 3.4 & 8.3 & 3.5\\
  \arrayrulecolor{black}
\rowcolor{gray!10} \multicolumn{20}{c}{\textit{\textbf{360M$_{\text{SmolLM}}$ (Length=2K)}}} \\
  & \raggedright BF16 & 100.0\% & 13.5 & 2.4 & 6.4 & 14.3 & 5.0 & 8.8 & 2.5 & 18.0 & 17.5 & 7.1 & 47.5 & 37.5 & 24.9 & 1.5 & 3.4 & 8.1 & 10.4 \\
  & \raggedright Int4$_{\text{HQQ}}$ & \multirow{2}{*}{-75.00\%} 
  & 13.4 & 2.7 & 6.1 & 14.1 & 5.5 & 8.4 & 3.0 & 16.2 & 15.4 & 11.2 & 47.5 & 37.5 & 23.4 & 1.3 & 3.7 & 9.0 & 10.1 \\
  & \raggedright Int4$_{\text{Quanto}}$ & & 13.3 & 2.4 & 6.2 & 13.7 & 5.4 & 8.7 & 2.6 & 15.4 & 17.4 & 7.3 & 47.5 & 37.3 & 24.4 & 1.0 & 3.7 & 8.4 & 11.0 \\
  & \raggedright Int2$_{\text{HQQ}}$ & \multirow{2}{*}{-87.50\%} & 10.8 & 2.7 & 4.7 & 8.3 & 5.4 & 5.9 & 1.9 & 9.9 & 10.0 & 8.4 & 45.2 & 27.5 & 14.2 & 2.1 & 4.2 & 10.0 & 11.9 \\
  & \raggedright Int2$_{\text{Quanto}}$ &  & 8.6 & 2.6 & 2.2 & 4.4 & 3.9 & 4.8 & 1.4 & 5.6 & 8.9 & 2.9 & 44.0 & 26.8 & 9.6 & 1.0 & 1.9 & 7.2 & 9.7 \\
  \hline
  \multirow{3}{*}{$32$} 
  & \raggedright BF16 & -68.75\% & 13.3 & 2.5 & 6.0 & 13.6 & 5.0 & 8.4 & 2.8 & 19.2 & 15.4 & 10.4 & 43.5 & 35.0 & 29.8 & 1.0 & 3.0 & 10.5 & 6.5 \\
  & \raggedright Int4$_{\text{HQQ}}$ & \multirow{2}{*}{-92.19\%}  & \bf 12.8 & 2.2 & 5.6 & 14.2 & 4.7 & 8.7 & 2.6 & 14.3 & 14.6 & 8.0 & 43.5 & 34.1 & 29.7 & 1.2 & 3.1 & 11.9 & 6.8 \\
  & \raggedright Int4$_{\text{Quanto}}$ & & \bf 12.9 & 2.1 & 5.2 & 11.9 & 5.0 & 8.9 & 2.7 & 15.9 & 15.1 & 10.6 & 43.5 & 31.8 & 27.0 & 0.7 & 3.0 & 16.6 & 7.1 \\
  \hline
  \multirow{3}{*}{$16$} 
  & \raggedright BF16 & -81.25\% & 10.9 & 2.0 & 5.5 & 13.5 & 4.9 & 9.9 & 3.1 & 13.1 & 13.8 & 10.7 & 26.5 & 27.0 & 19.7 & 0.8 & 4.0 & 13.0 & 6.3 \\
  & \raggedright Int4$_{\text{HQQ}}$ & \multirow{2}{*}{-95.31\%}  & 10.4 & 2.1 & 4.7 & 13.1 & 4.9 & 9.3 & 2.8 & 11.5 & 12.8 & 8.0 & 26.5 & 25.5 & 20.5 & 0.7 & 4.1 & 14.1 & 6.5 \\
  & \raggedright Int4$_{\text{Quanto}}$ &  & \bf 10.2 & 2.0 & 5.0 & 13.2 & 4.4 & 9.0 & 2.5 & 12.0 & 12.5 & 9.7 & 27.5 & 24.0 & 18.9 & 0.6 & 3.2 & 11.3 & 7.6 \\
  \hline
  \multirow{3}{*}{$8$} 
  & \raggedright BF16 & -87.50\% & 9.4 & 1.9 & 4.5 & 11.7 & 4.3 & 8.5 & 2.9 & 12.5 & 12.5 & 9.5 & 24.0 & 20.3 & 14.4 & 0.9 & 3.7 & 10.9 & 8.3 \\
  & \raggedright Int4$_{\text{HQQ}}$ & \multirow{2}{*}{-96.87\%} & 9.0 & 1.8 & 4.4 & 11.2 & 4.3 & 8.0 & 2.4 & 10.5 & 11.4 & 7.2 & 23.5 & 20.8 & 12.5 & 0.9 & 4.2 & 11.5 & 8.7\\
  & \raggedright Int4$_{\text{Quanto}}$ &  & 8.8 & 2.2 & 3.8 & 10.7 & 3.8 & 7.3 & 2.8 & 11.2 & 11.1 & 7.5 & 22.5 & 21.0 & 12.0 & 0.7 & 4.6 & 10.5 & 8.7\\
  \arrayrulecolor{black}
  \midrule

\end{tabular}
\caption{Evaluation results of all models on LongBench, including Task A: narrativeqa, B: qasper, C: multifieldqa\_en, D: hotpotqa, E: 2wikimqa, F: musique, G: gov\_report, H: qmsum, I: multi\_news, J: trec, K: triviaqa, L: samsum, M: passage\_count, N: passage\_retrieval\_en, O: lcc, P: repobench-p. \textbf{Bold} indicates compression ratios greater than or equal to Int2 quantization while also achieving performance higher than Int2.}
\label{tab:other_long_bench}
\end{table*}

\begin{table*}[t]
\centering
\small
\begin{tabular}{lcr@{\hspace{2pt}}lcccccc}
  \toprule
  \textbf{Model} & \textbf{Tokens} & \multicolumn{2}{l}{\textbf{Avg@CS}} & \textbf{MMLU} & \textbf{ARC} & \textbf{PIQA} & \textbf{HS} & \textbf{OBQA} & \textbf{WG}\\
  \midrule
  \rowcolor{gray!10}135M$_{\text{SmolLM}}$ & 600B & \multicolumn{2}{l}{44.50} & 29.80 & 42.43 & 68.06 & 41.09 & 33.60 & 52.01\\
  \textit{- full-rope} & \multirow{5}{*}{6B}  & 44.42 &  & 29.91 & 41.71 & 68.28 & 41.33 & 33.80 & 51.46 \\
  - $\mathcal{S}_{\text{high}}$ &   & 43.60&\textsubscript{-0.82} & 29.87 & 41.29 & 67.08 & 39.58 & 32.80 & 50.99 \\
  - $\mathcal{S}_{\text{low}}$ &   & 39.17&\textsubscript{-5.25} & 27.67 & 35.33 & 62.30 & 33.32 & 27.60 & 48.78 \\
  - $\mathcal{S}_{\text{uniform}}$ &   &\bf 44.01&\textsubscript{-0.41} & 29.79 & 41.09 & 67.95 & 40.54 & 34.20 & 50.51 \\
  - $\mathcal{S}_{\text{2-norm}}$ &   &  43.73&\textsubscript{-0.69} & 30.00 & 41.29 & 68.17 & 39.83 & 33.20 & 49.88 \\
  \arrayrulecolor{gray!20}
  \hline
  - $\mathcal{S}_{\text{high}}$ + SVD\textsubscript{joint} & \multirow{4}{*}{6B} & 40.85&\textsubscript{-3.57} & 28.46 & 37.25 & 64.85 & 35.31 & 30.20 & 49.01 \\
  - $\mathcal{S}_{\text{uniform}}$ + SVD\textsubscript{joint} &  & 41.79&\textsubscript{-2.63} & 28.74 & 39.30 & 65.83 & 36.37 & 31.20 & 49.33 \\
  - $\mathcal{S}_{\text{2-norm}}$ + SVD\textsubscript{joint} &   & \bf 42.18&\textsubscript{-2.24} & 28.79 & 40.11 & 65.94 & 36.68 & 31.20 & 50.36\\
  - $\mathcal{S}_{\text{2-norm}}$ + SVD\textsubscript{split} &  & 41.27 & \textsubscript{-3.15} & 28.05 & 38.65 & 65.51 & 34.04 & 31.20 & 49.17 \\
  \arrayrulecolor{black}
  \midrule
  \rowcolor{gray!10}1B7$_{\text{SmolLM}}$ & 1T & \multicolumn{2}{l}{55.90} & 39.27 & 59.87 & 75.73 & 62.93 & 42.80 & 54.85 \\
  \textit{- full-rope} & \multirow{5}{*}{6B}  & \multicolumn{2}{l}{55.71} & 38.66 & 59.02 & 75.79 & 62.60 & 43.20 & 55.01 \\
  - $\mathcal{S}_{\text{high}}$ &   & 54.80&\textsubscript{-0.91} & 38.18 & 57.57 & 75.08 & 60.66 & 42.40 & 54.93 \\
  - $\mathcal{S}_{\text{low}}$ &   & 53.84&\textsubscript{-1.87} & 37.49 & 55.24 & 74.16 & 59.22 & 42.60 & 54.30 \\
  - $\mathcal{S}_{\text{uniform}}$ &   &\bf 55.30&\textsubscript{-0.41} & 38.52 & 57.89 & 75.68 & 61.85 & 42.60 & 55.25 \\
  - $\mathcal{S}_{\text{2-norm}}$ &   &  54.98&\textsubscript{-0.73} & 38.33 & 57.47 & 76.06 & 61.77 & 41.40 & 54.85 \\
  \arrayrulecolor{gray!20}
  \hline
  - $\mathcal{S}_{\text{high}}$ + SVD\textsubscript{joint} & \multirow{4}{*}{6B}  & 54.17 &\textsubscript{-1.54} & 37.35 & 55.99 & 74.59 & 59.18 & 42.20 & 55.72 \\
  - $\mathcal{S}_{\text{uniform}}$ + SVD\textsubscript{joint} &   & 54.27&\textsubscript{-1.44} & 37.95 & 56.78 & 74.86 & 60.23 & 41.00 & 54.78 \\
  - $\mathcal{S}_{\text{2-norm}}$ + SVD\textsubscript{joint} &   & \bf 54.28&\textsubscript{-1.43} & 37.79 & 56.33 & 75.68 & 60.59 & 41.00 & 54.30 \\
  - $\mathcal{S}_{\text{2-norm}}$ + SVD\textsubscript{split} &   & 52.90  & \textsubscript{-2.81} & 36.99 & 53.80 & 73.39 & 58.55 & 41.60 & 53.04 \\
  \arrayrulecolor{black}
  \bottomrule
\end{tabular}
\caption{The complete results of the ablation experiment.}
\vspace{-0.4cm}
\label{tab:partial_rope_full}
\end{table*}
\begin{figure}[t]
  \centering
  \includegraphics[width=\linewidth]{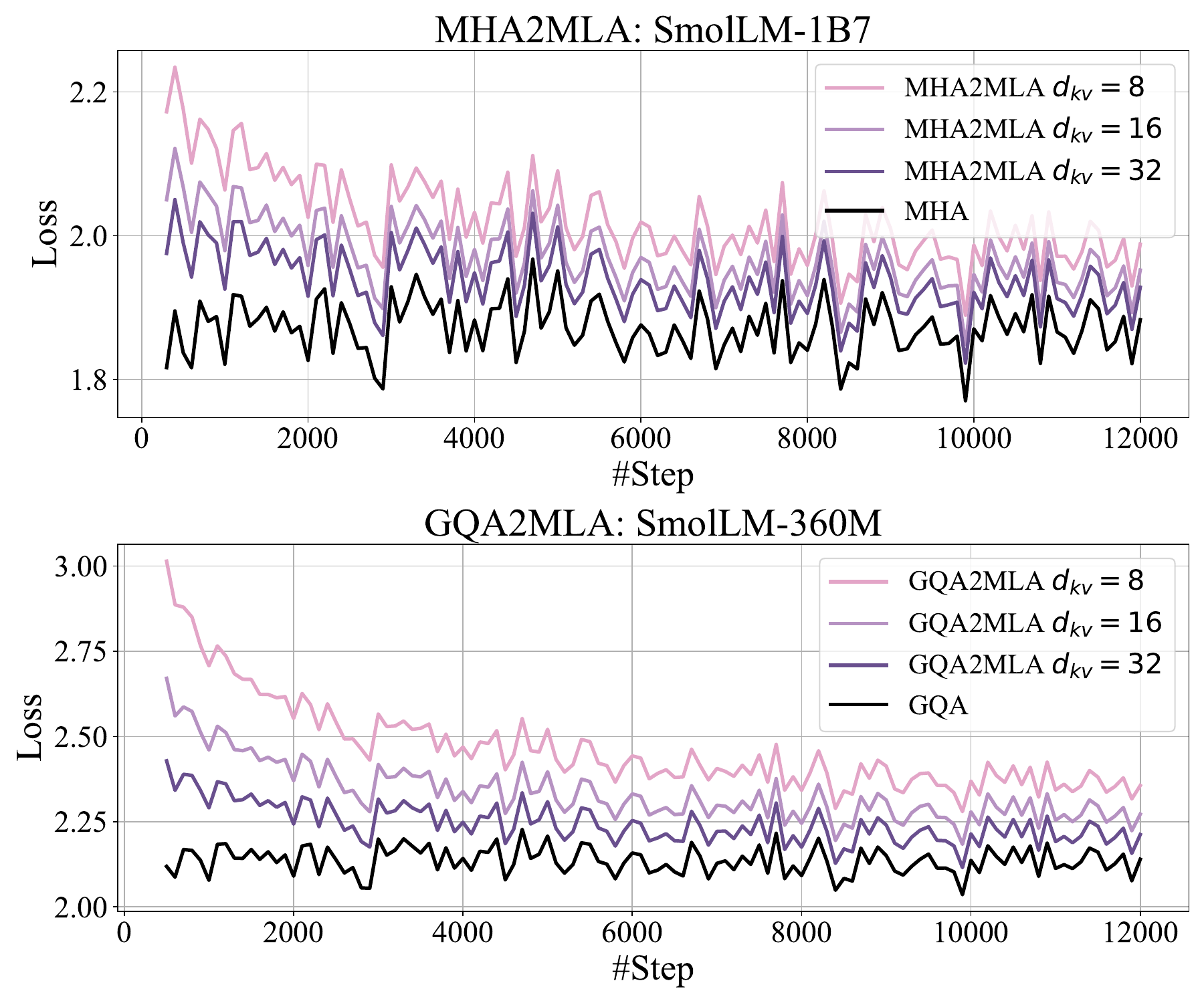}
  \caption{The fine-tuning loss curves under different KV cache storage ratios (with colors ranging from light to dark representing 12.5\%, 18.75\%, 31.25\%, and 100\%).}
  \label{fig:res_rank_loss}
  \vspace{-0.3cm}
\end{figure}
\begin{figure}[t]
  \centering
  \includegraphics[width=\linewidth]{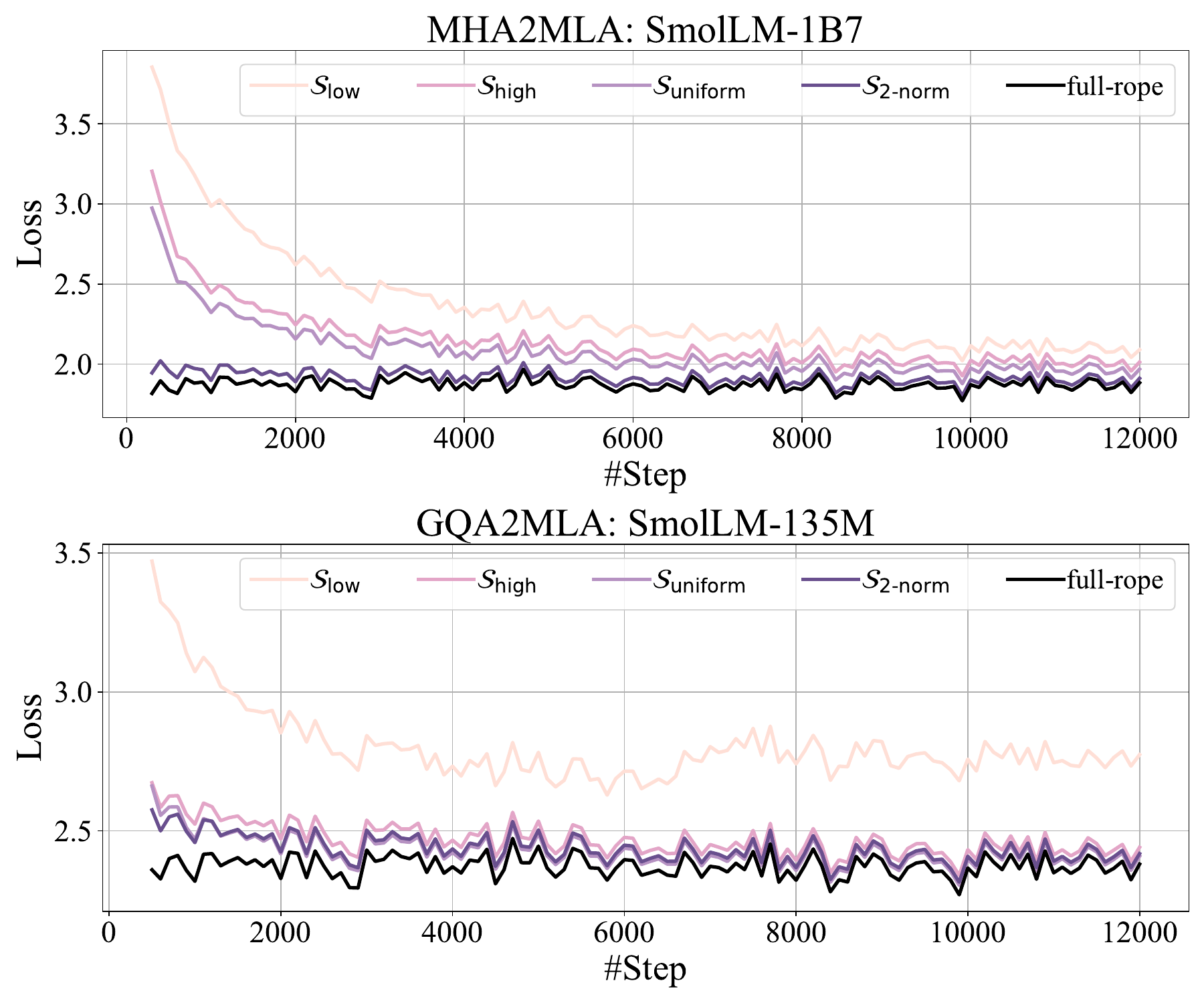}
  \caption{The fine-tuning loss curves under different partial-RoPE strategy.}
  \label{fig:pe_loss}
  \vspace{-0.3cm}
\end{figure}
\begin{figure}[t]
  \centering
  \includegraphics[width=\linewidth]{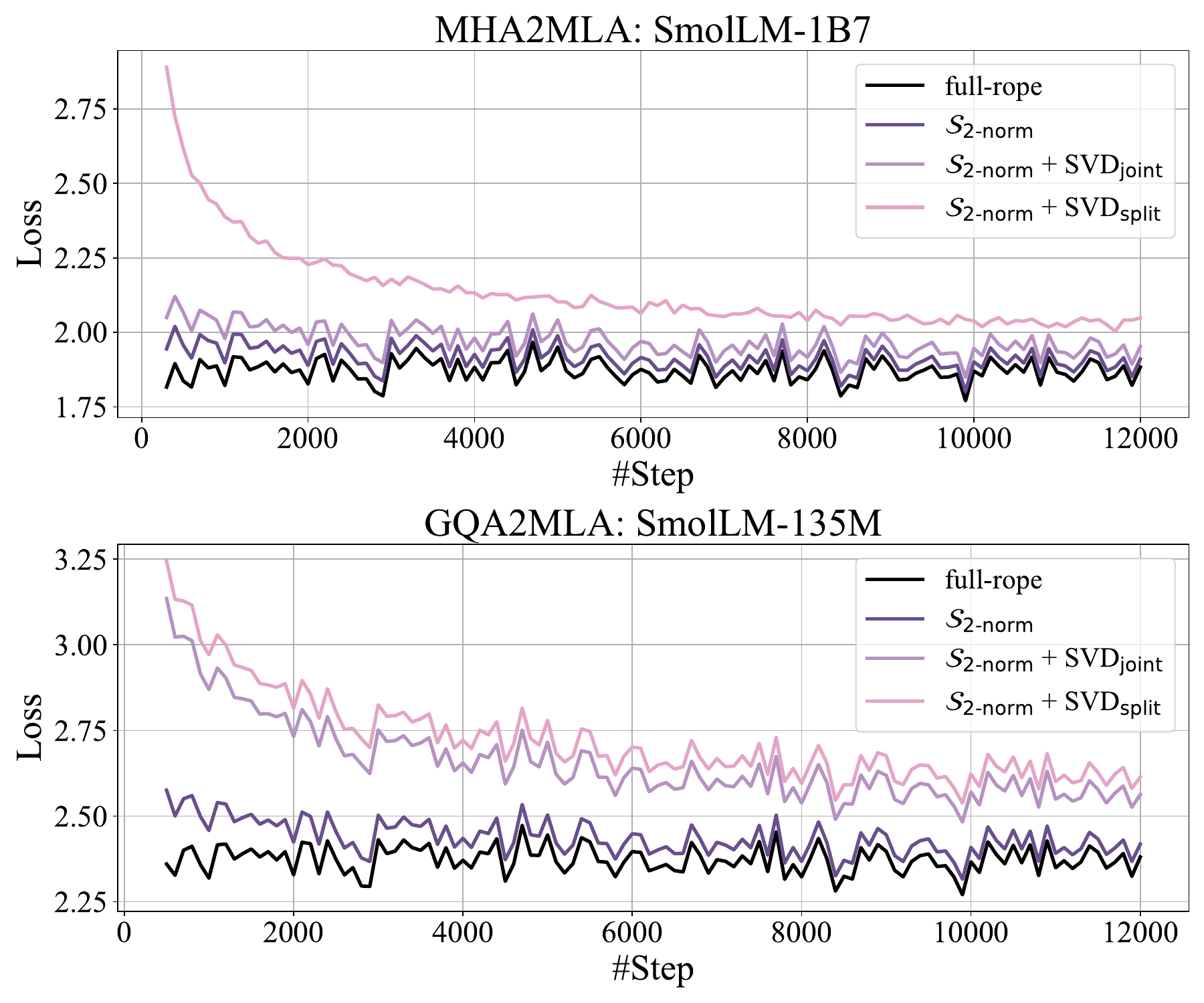}
  \caption{The fine-tuning loss curves under the combination of $\mathcal{S}_{\text{2-norm}}$ and different SVD strategies.}
  \label{fig:svd_loss}
  \vspace{-0.3cm}
\end{figure}


In this section, we present the detailed results. 
\paragraph{Detailed LongBench evaluation} is reported in \Cref{tab:other_long_bench}.

\paragraph{Detailed ablation experiment} is reported in \Cref{tab:partial_rope_full}.

\paragraph{Additional visualizations of fine-tuning loss} 
We present the loss of the other two models fine-tuned, excluding the ones mentioned in the main text, in \Cref{fig:res_rank_loss}. 
We observe that as fine-tuning progresses, the gap in loss between our approach and the baseline gradually decreases, and both exhibit similar fluctuations, demonstrating the effectiveness of our approach. 
In \Cref{fig:pe_loss}, we show the loss under different partial-RoPE strategies. Except for $\mathcal{S}_{\text{low}}$, the other three partial-RoPE schemes show little difference from the baseline. Additionally, $\mathcal{S}_{\text{low}}$ has a higher probability of convergence failure. In \Cref{fig:svd_loss}, we show the loss under different SVD strategies. The loss curves on both 1B7$_{\text{SmolLM}}$ and 135M$_{\text{SmolLM}}$ reveal that SVD\textsubscript{joint} outperforms SVD\textsubscript{split}.


\end{document}